\documentclass[a4paper, 10pt, conference]{ieeeconf}      

\IEEEoverridecommandlockouts                              
\usepackage{graphics} 
\usepackage{epsfig} 
\usepackage{mathptmx} 
\usepackage{times} 
\usepackage{amsmath} 
\usepackage{amssymb}  
\usepackage{cite}
\usepackage{algorithmic}
\usepackage{graphicx}
\usepackage{textcomp}
\usepackage{wrapfig}
\usepackage{multirow}
\graphicspath{{figures/}}

\usepackage[OT2,OT1]{fontenc} 
\usepackage{subfigure}

\usepackage[linkcolor=black,
            anchorcolor=black,
            citecolor=black,
            urlcolor =black,
            hidelinks]{hyperref}

\title{\LARGE \bf
Prototype Design and Efficiency Analysis of a Novel Robot Drive 

Based on 3K-H-V Topology
}

\author{Le Qi, Dapeng Yang, Baoshi Cao$^{*}$, Zhiqi Li, Yikun Gu, Zongwu Xie, and Hong Liu
\thanks{This work is partially supported by China Postdoctoral Science Foundation (Grant NO.2021M690826), and supported by Self-Planned Task (NO.SKLRS202101A04) of State Key Laboratory of Robotics and System (HIT) }
\thanks{Le Qi, Dapeng Yang, Baoshi Cao, Zhiqi Li, Yikun Gu, Zongwu Xie, and Hong Liu are with the State Key Laboratory of Robotics and System, Harbin Institute of Technology, Harbin 150080, China (e-mail: qile@hit.edu.cn; yangdapeng@hit.edu.cn; cbs@hit.edu.cn; lzq@hit.edu.cn; guyikun@hit.edu.cn; xiezongwu@hit.edu.cn; hong.liu@hit.edu.cn).}
\thanks{$^{*}$Corresponding Author: Baoshi Cao (cbs@hit.edu.cn).}}

\begin{document}
\maketitle
\thispagestyle{empty}
\pagestyle{empty}
\begin{abstract}
    Robot actuators directly affect the performance of robots, and robot drives directly affect the performance of robot actuators. With the development of robotics, robots have put higher requirements on robot drives, such as high stiffness, high accuracy, high loading, high efficiency, low backlash, compact size, and hollow structure. In order to meet the demand development of robot actuators, this research base proposes a new robot drive based on 3K-H-V topology using involute and cycloidal gear shapes, planetary cycloidal drive, from the perspective of drive topology and through the design idea of decoupling. In this study, the reduction ratio and the efficiency model of the 3K-H-V topology were analyzed, and a prototype planetary cycloidal actuator was designed. The feasibility of the drive is initially verified by experimentally concluding that the PCA has a hollow structure, compact size, and high torque density (69 Nm/kg).
\end{abstract}

\section{INTRODUCTION}

Robot joints are usually driven by motor actuators. The critical components of a motor actuator are the motor and the drive. The performance of robot drives directly affects the performance of actuators and robots. Robot drives are not ideal torque amplifiers, which can bring adverse effects such as backlash, friction, and flexibility to the actuator\cite{schempf1993study}.

Harmonic drives, cycloidal drives, and RV drives with large reduction ratios are the common forms of robot drives\cite{pham2018high}. The cycloidal drive and RV drive are mainly used for large load applications, and the RV can achieve a high reduction ratio based on the cycloidal drive. However, the RV drive cannot be designed as a hollow structure with a high reduction ratio. Harmonic drives have been the main drives in robotics applications, especially in lightweight robots such as cobots \cite{rader2017highly}, but usually have low stiffness and poor efficiency \cite{matsuki2019bilateral}.

Some studies focus on the quasi-direct drive with low reduction ratios. Such designs usually have high backdrivability but must be matched with high-torque motors, primarily used in legged robots \cite{bledt2018mit}. The compound planetary type drive is also a new research direction, which can have extremely high reduction ratios in compact sizes \cite{garcia2020compact}, such as Wolfrom drives \cite{kapelevich2014gear} and Dojen drives \cite{introducing}. However, compound planetary drives have significant cyclic power losses, which means it is not easy to achieve high efficiency at high reduction ratios \cite{sensinger2013efficiency, chen2007virtualpower}. Some studies improve the drive efficiency by optimizing the design\cite{{akiyama2019highly},{matsuki2019bilateral}}, or further improve the structure to optimize the drive efficiency\cite{garcia2022r2power}. 

Robots are increasingly being used in production life. New applications are placing higher demands on robot performance and, in turn, robot actuators and, particularly, robot drives. In order to improve the disadvantages of existing robot drives in different aspects and to ensure the requirements of high torque density, high accuracy, high stiffness, high reduction ratio, high efficiency, low backlash, and hollow structure in a comprehensive way, this study proposes a new robot drive, planetary cycloidal drive, based on the basic principle of planetary differential drive and using the design idea of decoupling. The planetary cycloidal drive is based on the 3K-H-V topology with two tooth shapes, involute and cycloidal, and has good accuracy, stiffness, and efficiency performance, and can achieve a high reduction ratio and hollow structure. Based on this drive prototype, this study derives its forward and reverse efficiency equation based on the meshing efficiency, designs a robot actuator prototype, and performs efficiency and loading experiments.

\begin{table*}[!t]
	\centering
    \caption{The topology of common robot drive}
    \label{table:topo}
    \newcommand{\tabfig}[2]
    {   
        \multirow{5}{*}{\begin{tabular}[l]{@{}l@{}}
            \includegraphics[height=0.17\columnwidth]{#1} \includegraphics[height=0.17\columnwidth]{#2}
        \end{tabular}}
        
    } 

    \setlength{\tabcolsep}{3pt}
    \begin{tabular}{lcccccc}
    \hline
    & \multicolumn{1}{c}{2K-H} & \multicolumn{1}{c}{3K} & \multicolumn{1}{c}{\begin{tabular}[c]{@{}c@{}}2K-H NN$^{*}$\end{tabular}} & \multicolumn{1}{c}{K-H-V} & \multicolumn{1}{c}{2K-V} & \multicolumn{1}{c}{Harmonic} \\\hline
    \multirow{5}{*}{\begin{tabular}[l]{@{}l@{}}Type\end{tabular}} 
    &\tabfig{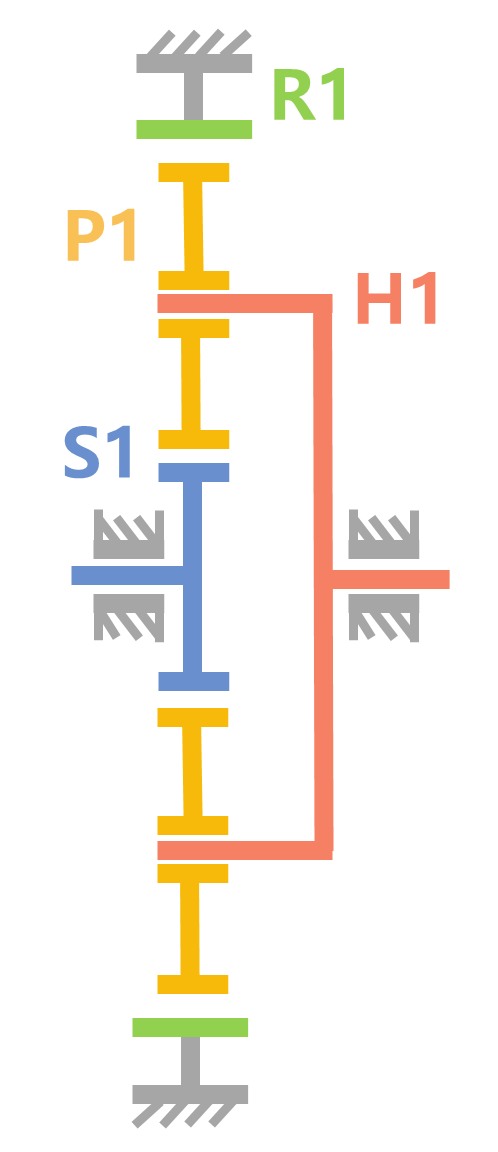}{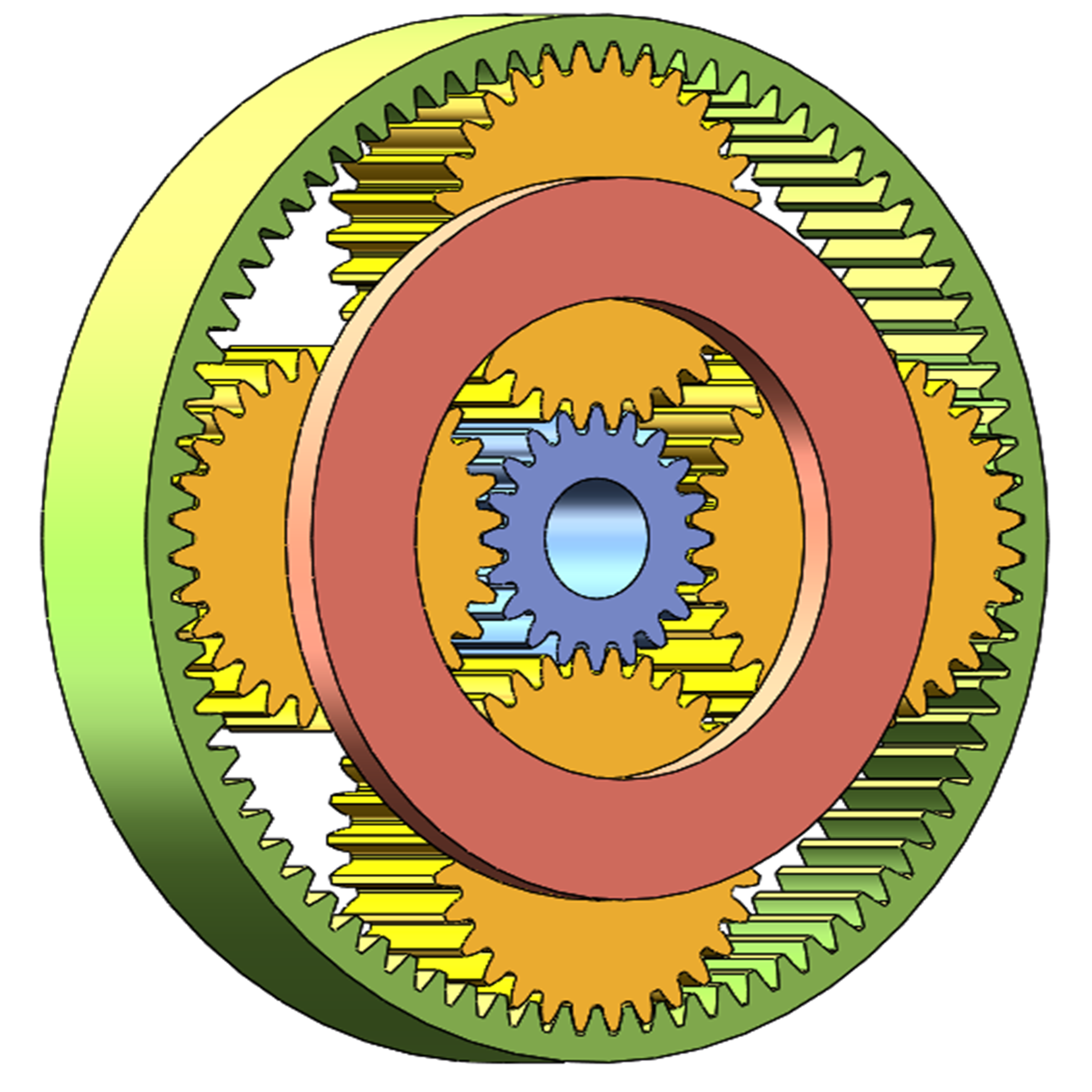} 
    &\tabfig{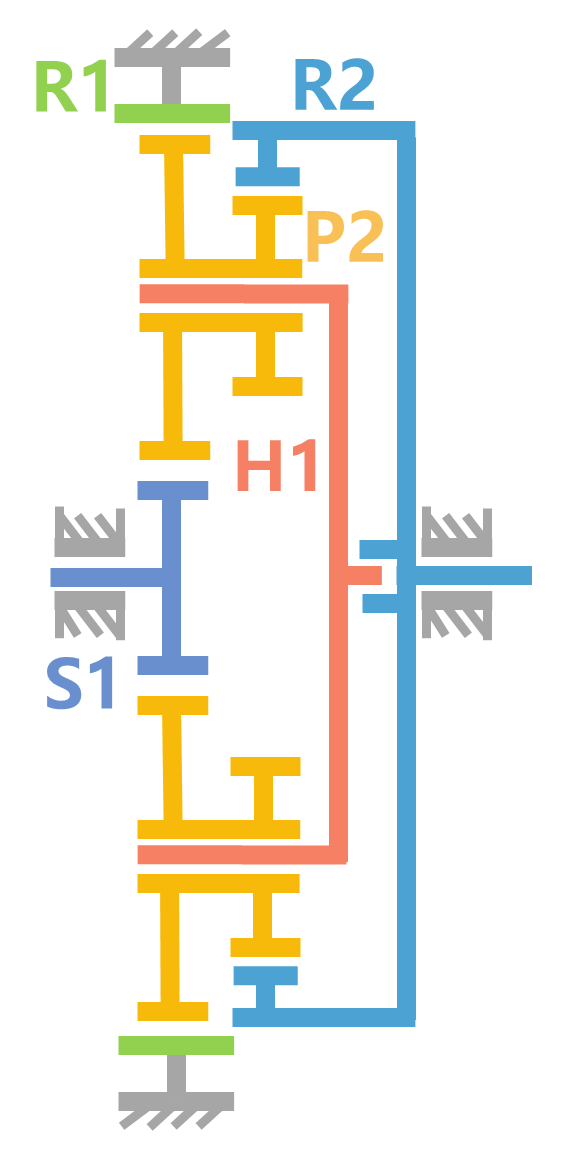}{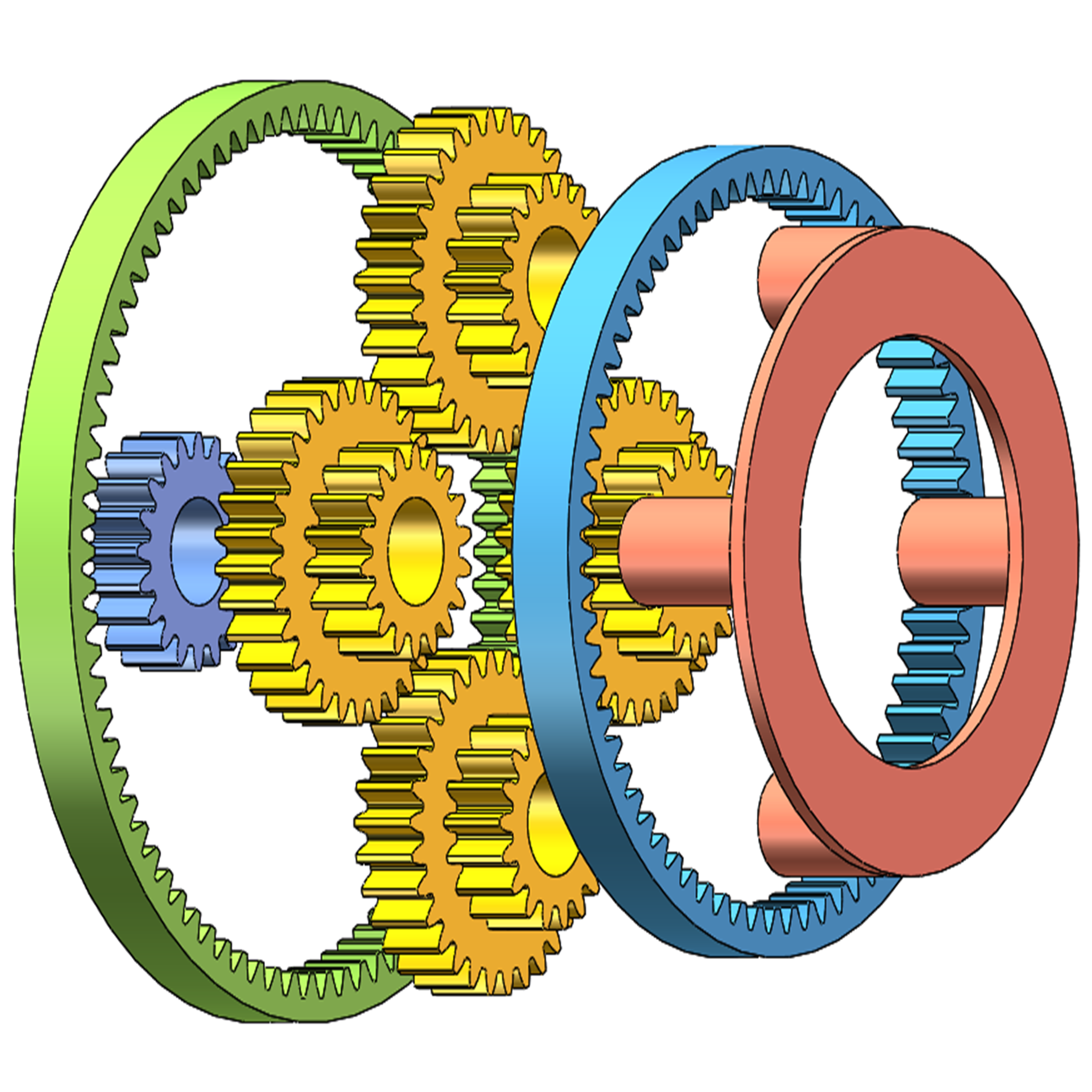} 
    &\tabfig{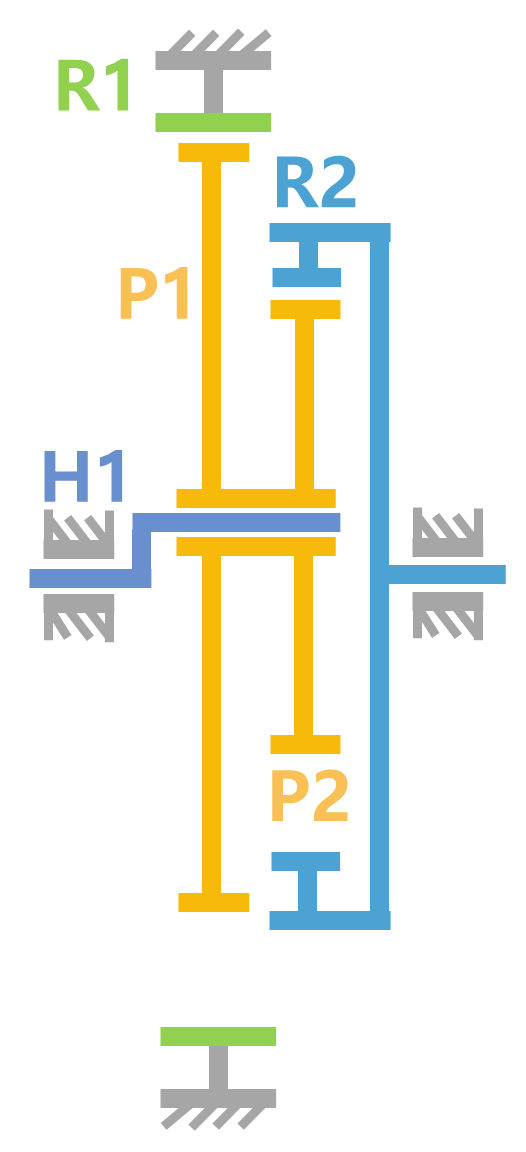}{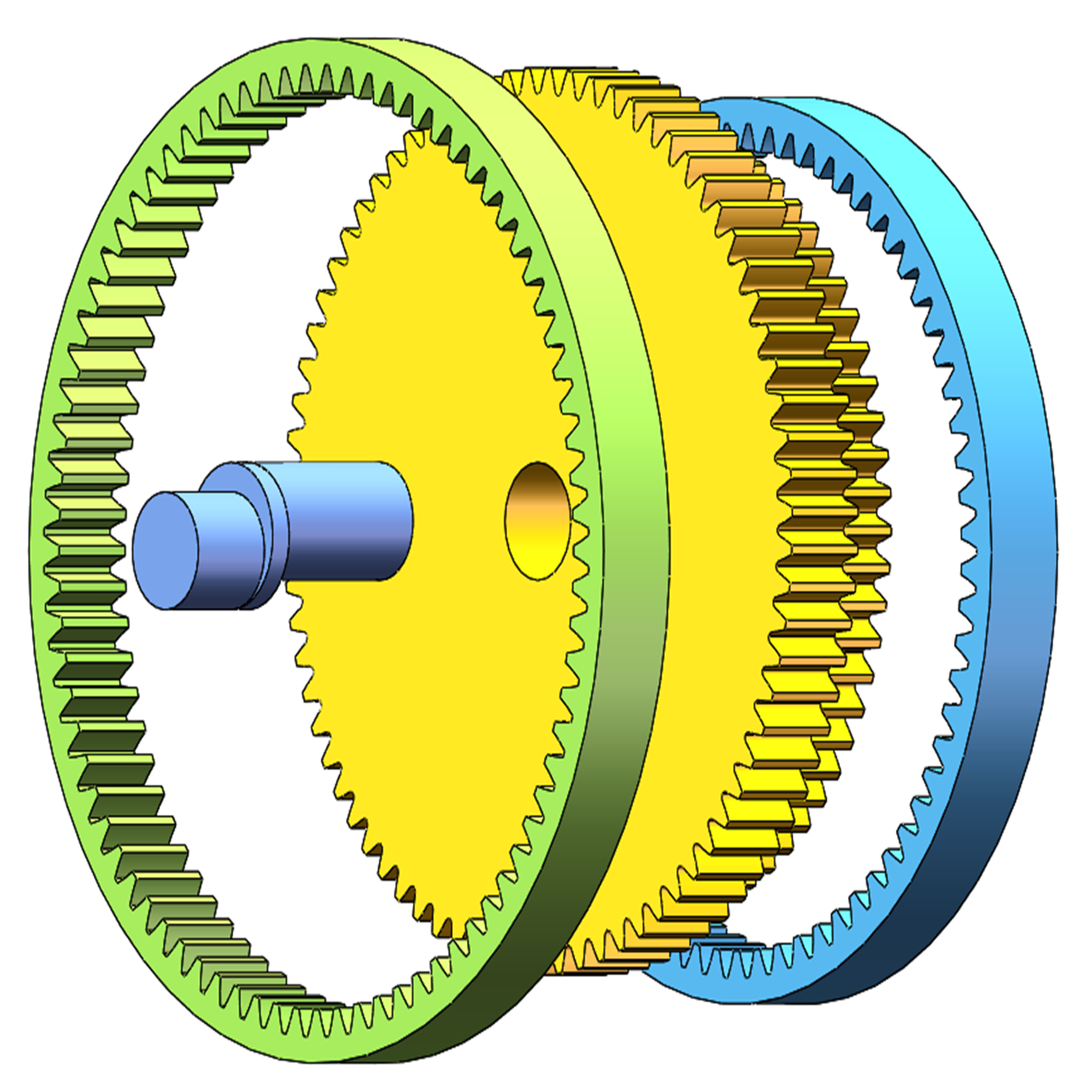}
    &\tabfig{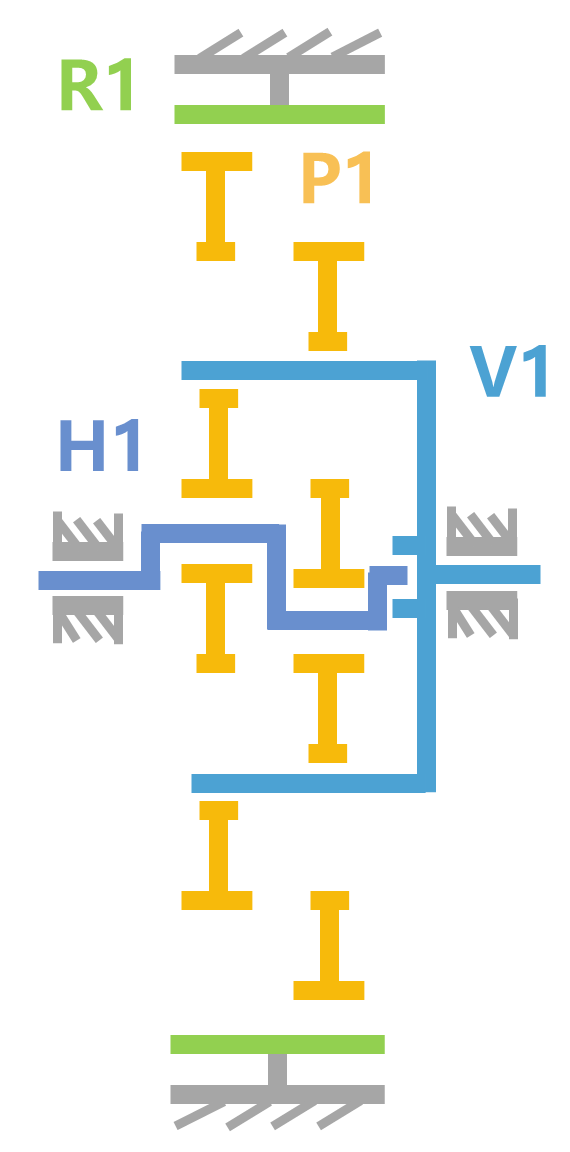}{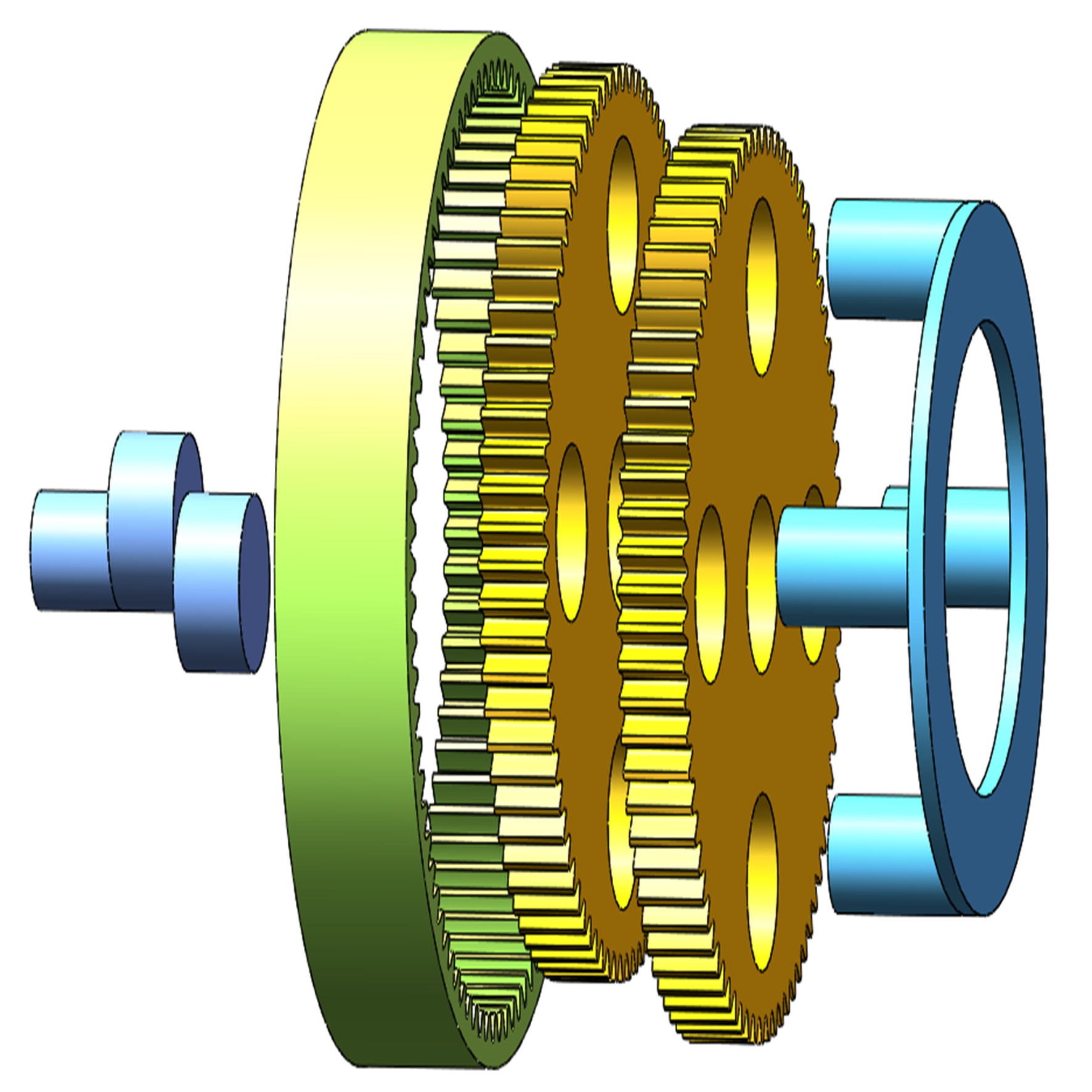} 
    &\tabfig{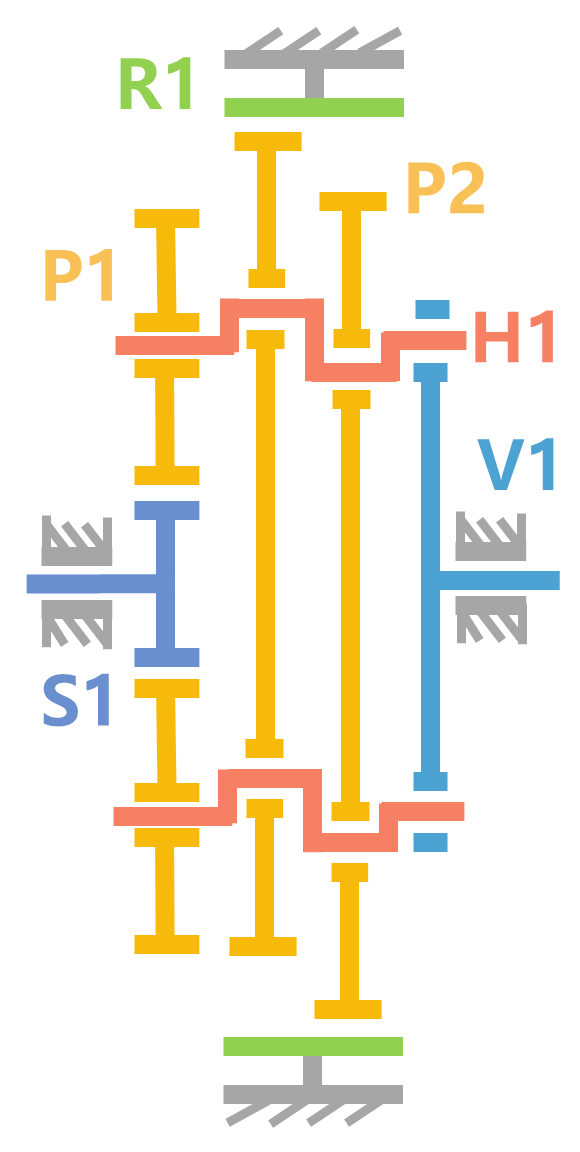}{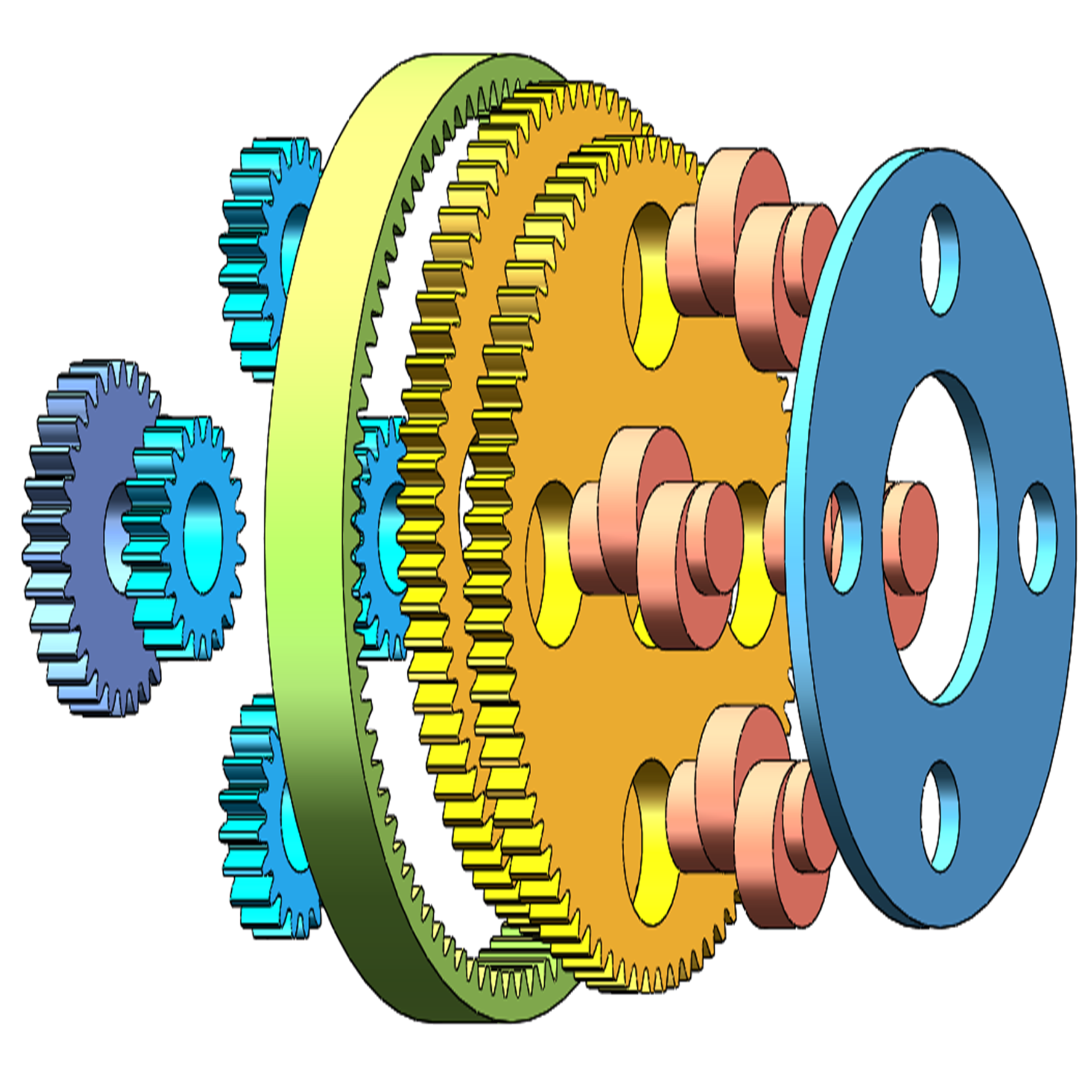} 
    &\tabfig{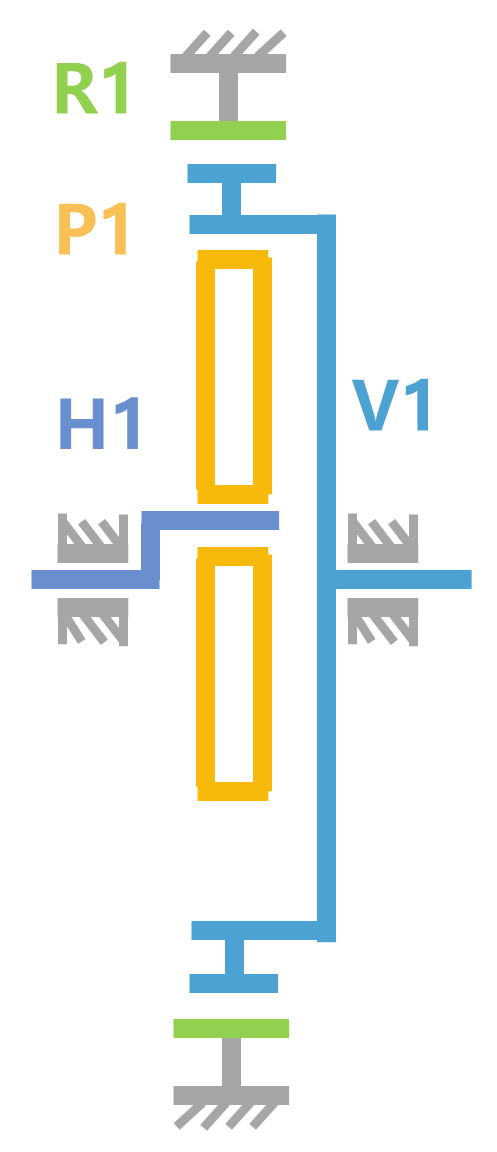}{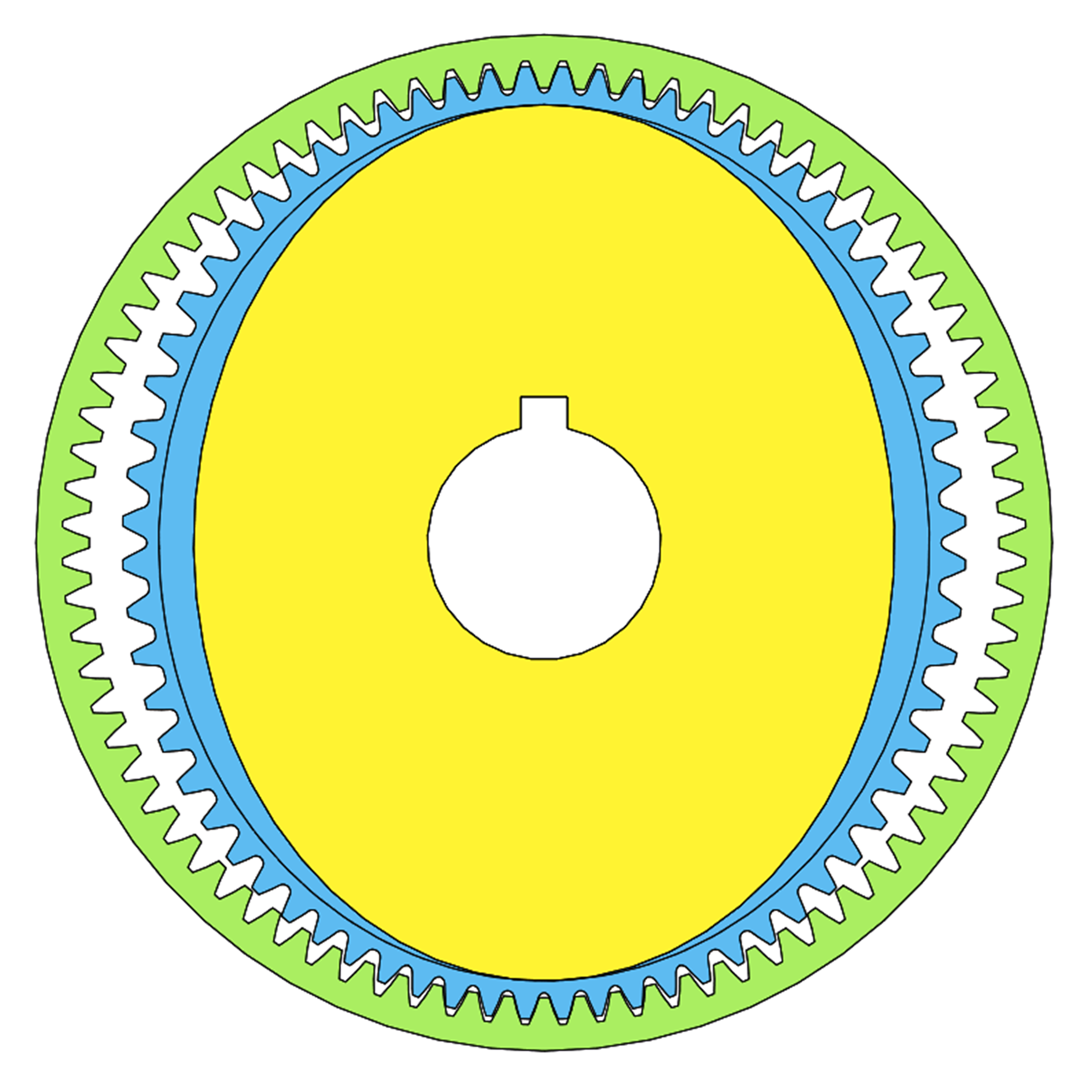} 
    \\
    \\
    \\
    \\
    \\\hline
    Drive &
    \multicolumn{1}{c}{\begin{tabular}[c]{@{}c@{}}Planetary\end{tabular}} &
    \multicolumn{1}{c}{\begin{tabular}[c]{@{}c@{}}Wolfrom\end{tabular}} &
    \multicolumn{1}{c}{\begin{tabular}[c]{@{}c@{}}Dojen\end{tabular}} &
    \multicolumn{1}{c}{\begin{tabular}[c]{@{}c@{}}Cycloid\end{tabular}} &
    \multicolumn{1}{c}{\begin{tabular}[c]{@{}c@{}}RV \end{tabular}} &
    \multicolumn{1}{c}{\begin{tabular}[c]{@{}c@{}}Harmonic\end{tabular}} \\ \hline
    Reduction ratio & +     & ++++ & ++++ & ++   & +++ & +++   \\ \hline
    Efficiency  & ++++  & +    & +    & +++  & +++ & ++    \\ \hline
    \multicolumn{7}{l}{$^{*}$ NN is used to distinguish the compound 2K-H from the common 2K-H, indicating that there are two internal gear meshes.}                               
    \end{tabular}
\end{table*}

\section{ROBOT DRIVE BASED ON 3K-H-V TOPOLOGY}
The performance of robot drive is directly related to topology and teeth design. We follow the method proposed by 
BH. (K denotes the central gear, H denotes the planetary carrier, and V denotes the output mechanism) to classify common robot drives, as shown in Table \ref{table:topo}. The drive type known as planetary drive belongs to the 2K-H of them. The different topologies have principle differences in reduction ratios and efficiency characteristics. Since the specific teeth shape and number of teeth also affect the drive performance, the reduction ratio and efficiency parameters in Table \ref{table:topo} represent only a qualitative description of the ease of achieving a high reduction ratio or high efficiency in that topology. Where the more "+" means that the structure is easier to achieve the corresponding performance \cite{sensinger2013efficiency}. In robot drives, the involute gear is the most common, which has a high meshing efficiency, but usually has a large backlash and less number of teeth meshed. The cycloidal gear and pin-wheel are usually used in K-H-V and 2K-V to achieve high reduction ratios, low backlash, and high loading properties, while changing sliding friction into rolling friction to improve meshing efficiency. In harmonic drives, special teeth profiles such as "S" teeth achieve zero backlash performance using flexible structures.

\begin{figure}[!htbp]
	\centering
	\includegraphics[width=0.7 \columnwidth]{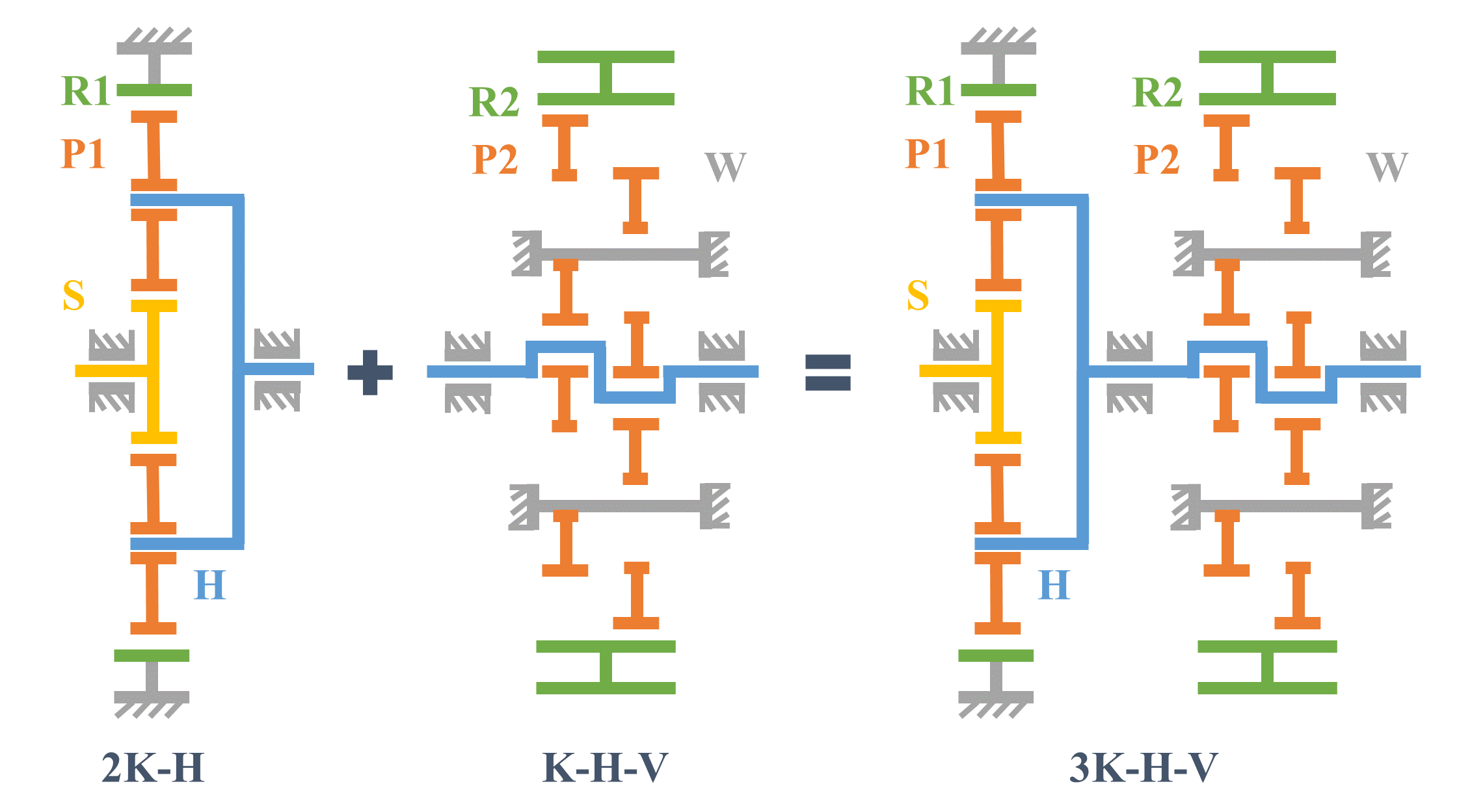}
	\caption{The 3K-H-V topology}
	\label{fig:3khv-0}
\end{figure}

In order to better meet the requirements of the robot drive, especially to deal with the conflict between high reduction ratio and high efficiency, this study proposes to decouple the input and output structures. By the decoupling design, suitable topology and teeth shape can be selected in the input and output stages, according to the demand, and the two stages can be optimized separately to reduce the parameter coupling. Through this decoupling design idea, a planetary cycloidal drive based on the 3K-H-V topology was proposed in this study, as shown in Table \ref{table:topo2}. The 3K-H-V topology is shown in Fig. \ref{fig:3khv-0}.

\begin{table}[!h]
    \centering
    \newcommand{\tabfig}[1]
    {   
        \multirow{9}{*}{\begin{tabular}[l]{@{}l@{}}
            \includegraphics[height=0.30\columnwidth]{#1}
        \end{tabular}}
        
    } 
    \caption{The planetary cycloid drive based on 3K-H-V }
    \label{table:topo2}
    \begin{tabular}{lcc}
        \hline
                      & Input stage    & Output stage\\ \hline
        Condition     & \begin{tabular}[c]{@{}c@{}}High speed\\ Low load\end{tabular} 
                      & \begin{tabular}[c]{@{}c@{}}Low speed\\ High load\end{tabular} \\ \hline
        Topology      & 2K-H           & K-H-V\\ \hline
        Input         & Sun gear       & Carrier\\ \hline
        Output        & Carrier        & Ring gear\\ \hline
        Tooth profile & Involute gears & \begin{tabular}[c]{@{}c@{}}Cycloid gears
                                        \\ Pin-wheels\end{tabular} \\ \hline
        \multirow{9}{*}{\begin{tabular}[l]{@{}l@{}}3D model \end{tabular}} 
                      &\tabfig{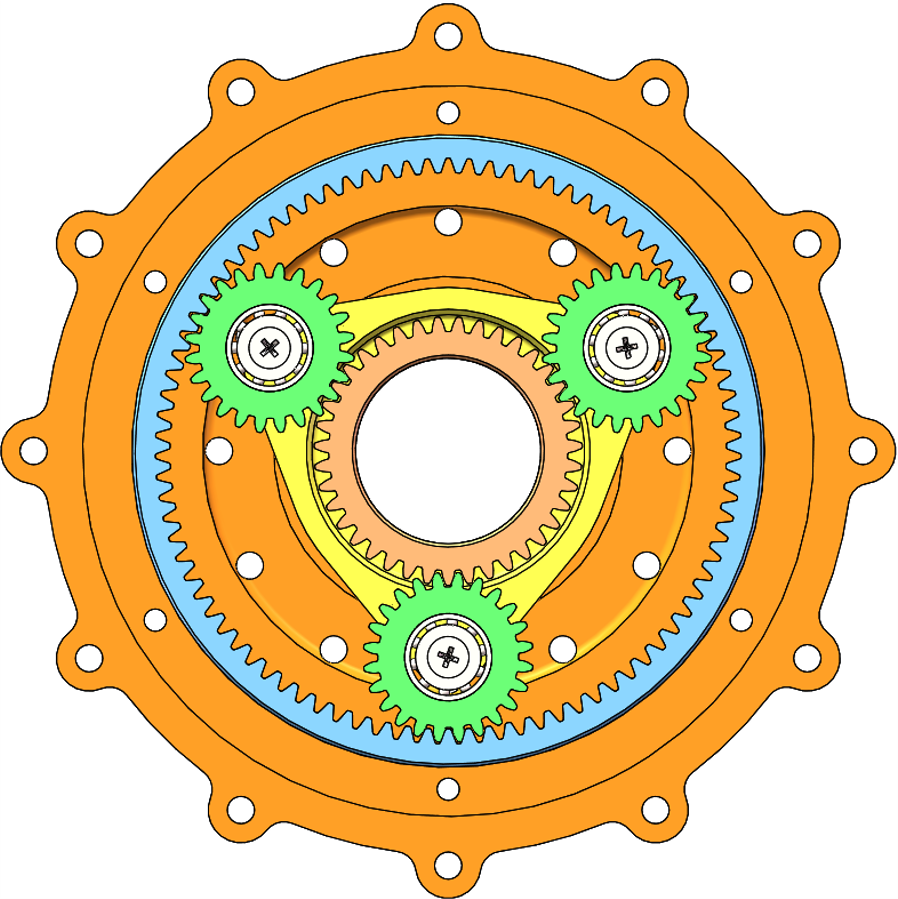} 
                      &\tabfig{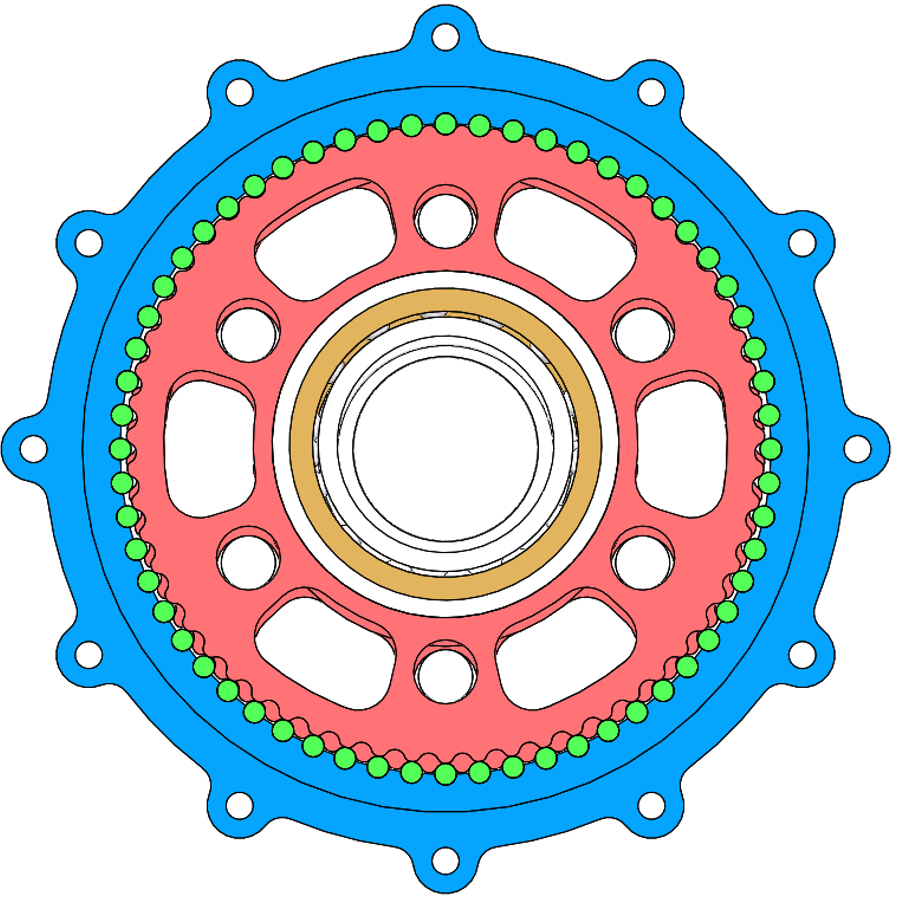} 
                      \\
                      \\
                      \\
                      \\
                      \\
                      \\
                      \\
                      \\
                      \\ \hline
        Property      & \begin{tabular}[c]{@{}c@{}}High speed\\ High efficiency\\ Hollow structure
                        \end{tabular}     
                      & \begin{tabular}[c]{@{}c@{}}High accuracy\\ High stiffness\\ High efficiency\\ High loading\\ Low backlash\\ Hollow structure
                        \end{tabular} \\ \hline
        \end{tabular}
\end{table}

\begin{figure}[!htbp]
	\centering
	\includegraphics[width=\columnwidth]{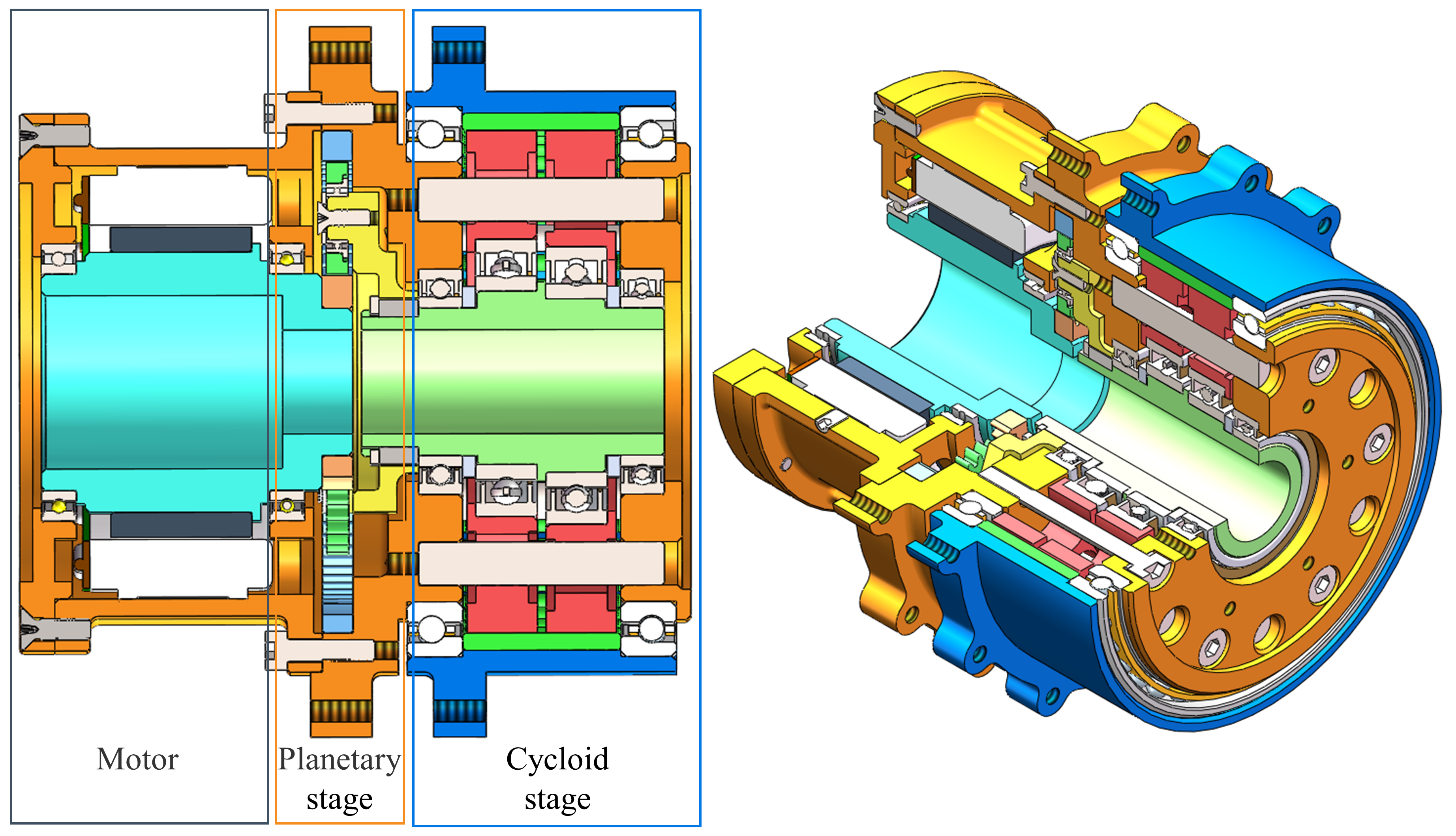}
	\caption{The planetary cycloid drive with motor}
	\label{fig:pca}
\end{figure}

The planetary cycloidal drive (PCD) obtains high accuracy, high stiffness, and high loading properties by using cycloidal K-H-V in the output stage, avoiding the flexibility problem of the harmonic drive. And PCD obtains both high efficiency and high reduction ratio properties by using involute 2K-H in the input stage. The RV drive also has a similar design idea, but the two-stage geometric parameters in the RV drive are highly coupled and cannot be designed as a hollow structure with a high reduction ratio. The PCD solves that problem, the low coupling of the two-stage geometric parameters facilitates the optimal design, and it is easy to get a hollow structure, which is excellent for application in robot joint actuators. Based on the PCD, a planetary cycloidal actuator (PCA) has been initially designed in this study, and its structural model is shown in Fig. \ref{fig:pca}.

\section{EFFICIENCY ANALYSIS OF 3K-H-V TOPOLOGY}
An important purpose of the PCD using 3K-H-V is to deal with the conflict between a high reduction ratio and high efficiency at the topology level through a two-stage decoupling design. This section analyzes the relationship between the number of teeth, kinematics, forward efficiency, and backward efficiency of each level of 3K-H-V to provide a base for further optimization of the PCD design. Only the teeth meshing efficiency loss, which accounts for the largest proportion of PCD efficiency, is considered in the efficiency loss, and the efficiency loss in other aspects is ignored.

\subsection{2K-H topology}

\subsubsection{Kinematics of 2K-H topology}

\begin{figure}[!t]
    \centering
    \subfigure[]{\includegraphics[width=0.48 \columnwidth]{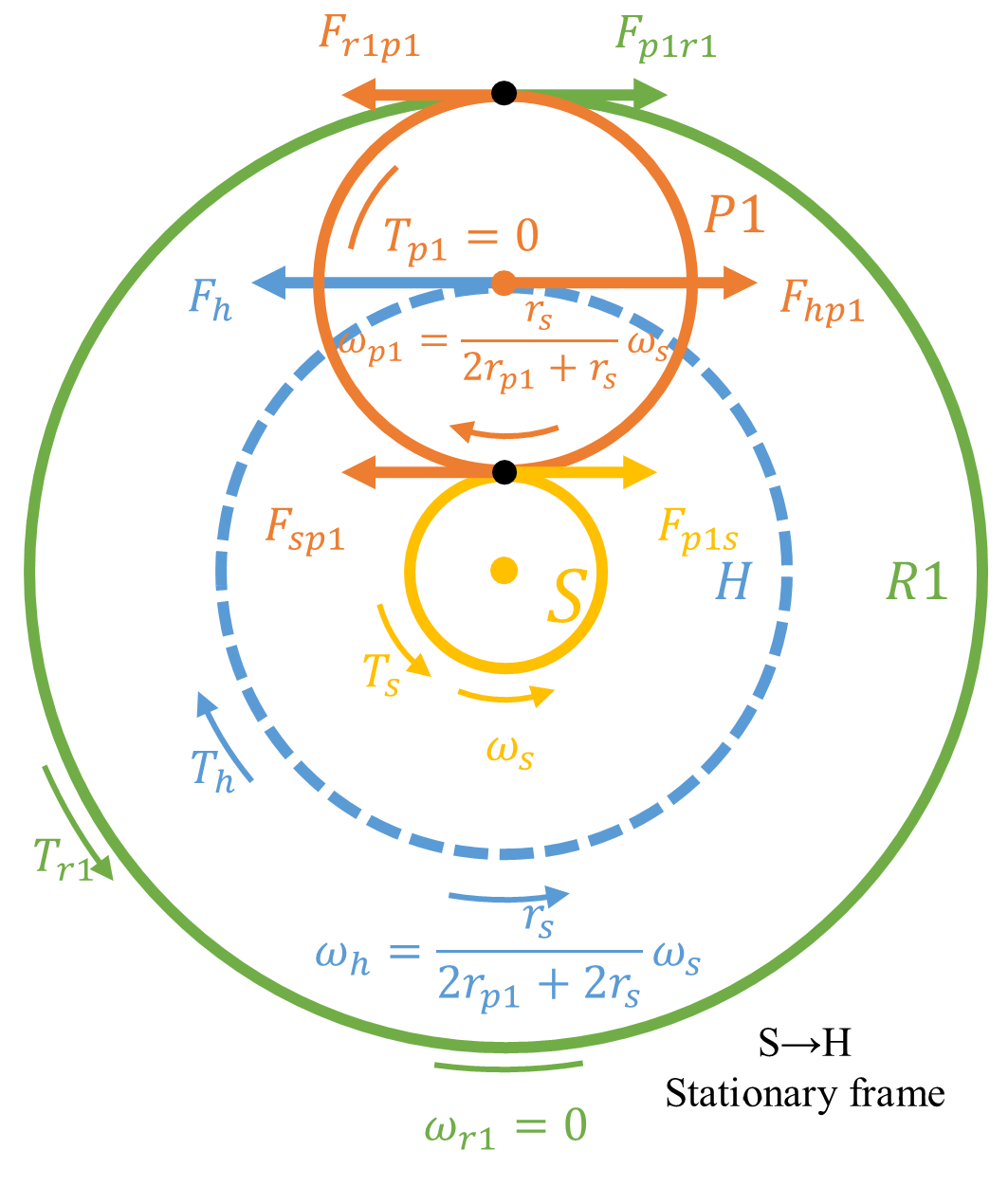}\label{fig:2K-H-1}}%
    \hfil
    \subfigure[]{\includegraphics[width=0.48 \columnwidth]{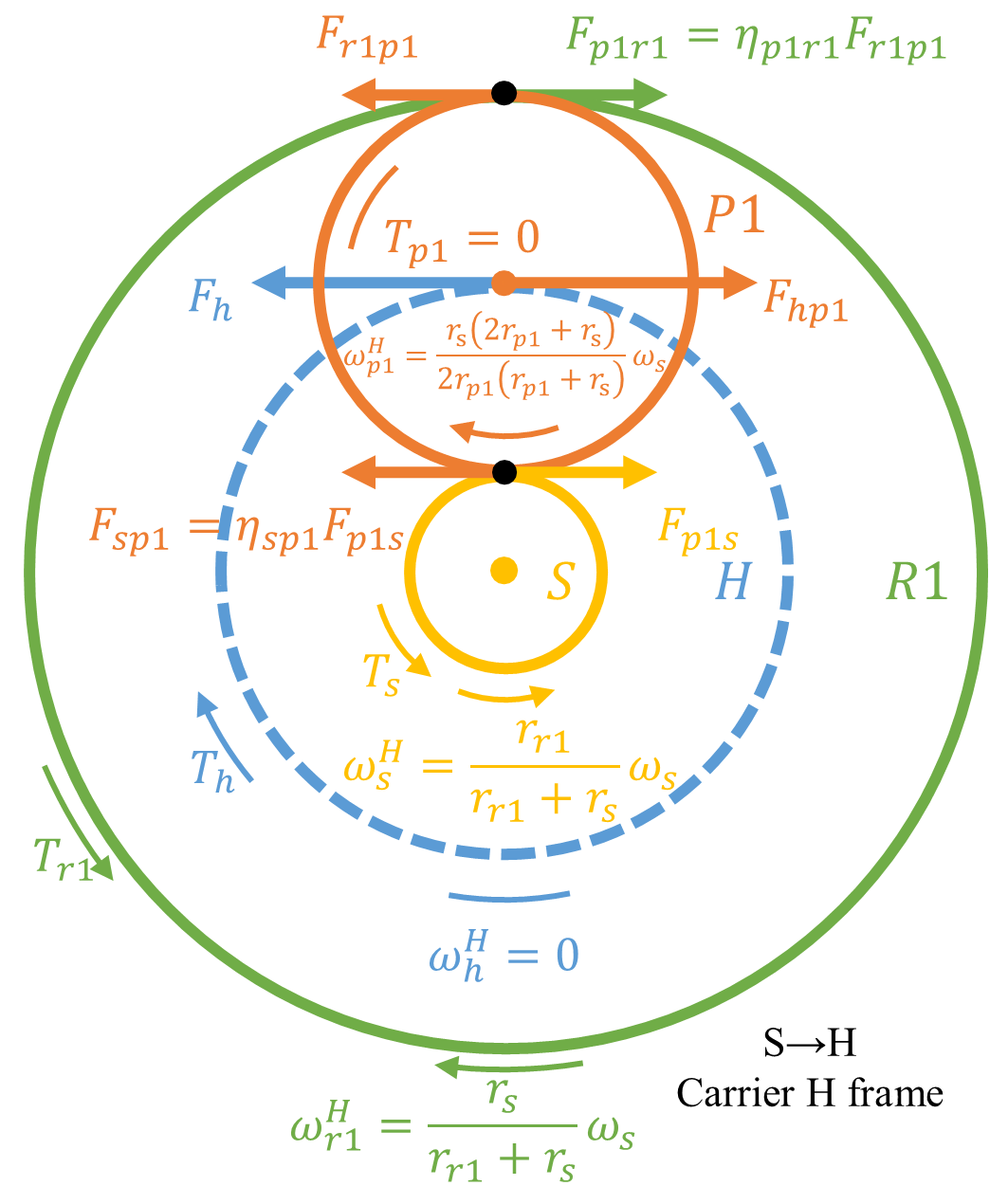}\label{fig:2K-H-2}}%
    \\
    \subfigure[]{\includegraphics[width=0.48 \columnwidth]{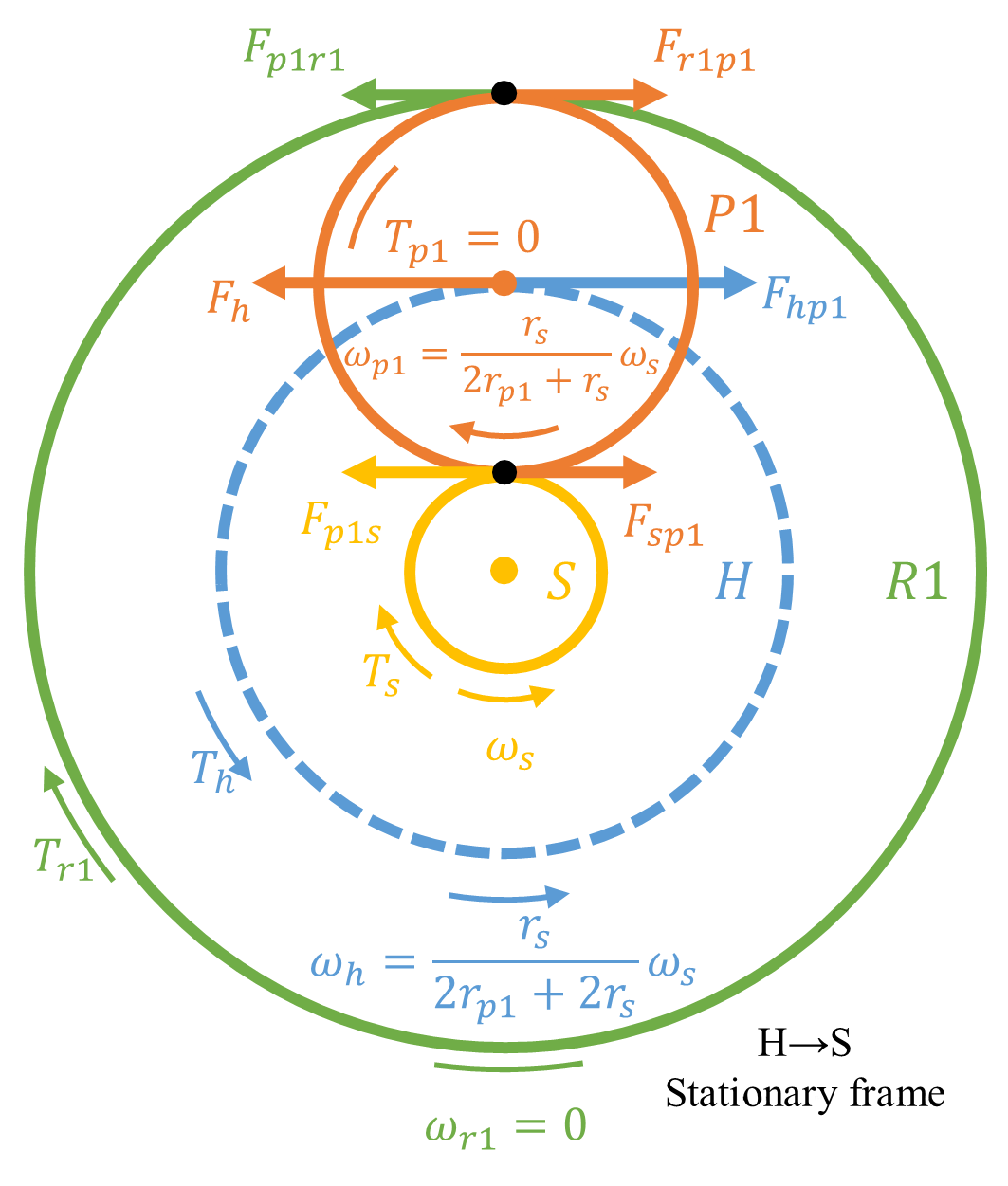}\label{fig:2K-H-3}}%
    \hfil
    \subfigure[]{\includegraphics[width=0.48 \columnwidth]{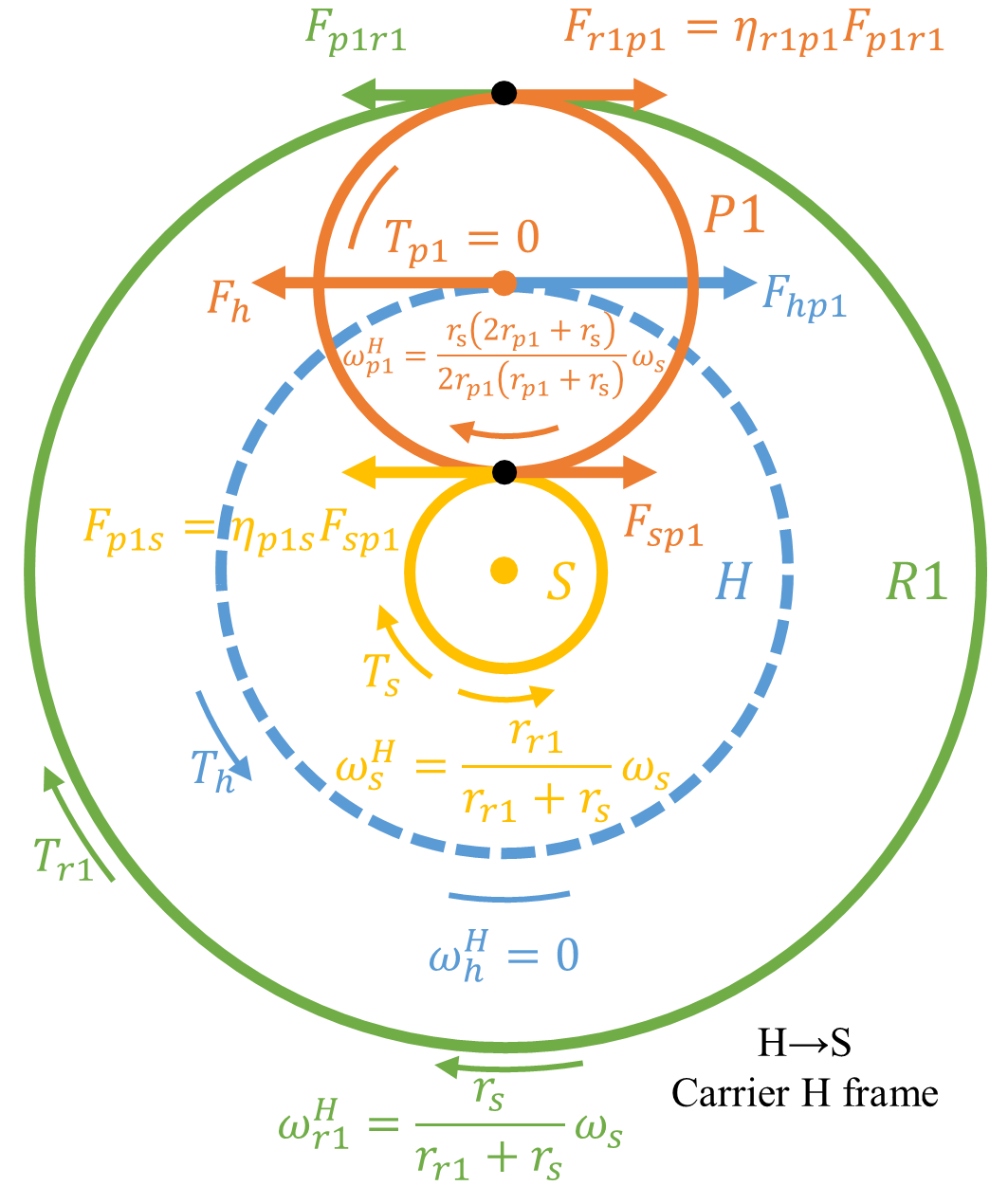}\label{fig:2K-H-4}}%
    \caption{The motion of each component of the 2K-H topology in different conditions.}
    \label{fig:2K-H}
    
\end{figure}

The 2K-H topology comprises two central gears (sun gear S and internal gear R1), a planetary carrier H, and planetary gears P1. Fig. \ref{fig:2K-H} shows the motion of the components in 2K-H in the stationary coordinate system and H coordinate system, respectively. The drive from S to H is the forward drive, and the drive from H to S is the backward drive.

In 2K-H, the pitch circle radiuses of S, R1, P1 and H are $r_{s}$, $r_{r1}$, $r_{p1}$ and $r_{h1}$, the teeth numbers of S, R1, and P1 are $z_{s}$, $z_{r1}$ and $z_{p1}$. The pitch circle radiuses and teeth numbers are in the following relations.
\begin{equation}
    \left\{\begin{aligned}
        \dfrac{r_s}{z_s} &= \dfrac{r_{p1}}{z_{p1}}  = \dfrac{r_{r1}}{z_{r1}}\\
        r_{r1} &= r_s+2r_{p1}\\
        r_{h1} &= r_s+r_{p1}
    \end{aligned}\right.
    \label{eq:eq1}
\end{equation}

As shown in Fig. \ref{fig:2K-H}, in the stationary coordinate system, the absolute speed of S, R1, P1 and H are $\omega_s$, $\omega_{r1}$, $\omega_{p1}$ and $\omega_{h}$, respectively. In the coordinate system of H, the relative speeds of S, R1, P1 and H are $\omega_{s}^H$, $\omega_{r1}^H$, $\omega_{p1}^H$ and $\omega_{h}^H$, respectively. The absolute speeds and relative speeds are in the following relationship. The negative sign means the opposite direction.
\begin{equation}
    \begin{array}{ll}
        \left\{\begin{array}{l}
            \omega_{p 1}=\dfrac{-r_{s}}{2 r_{p 1}+r_{s}} \omega_{s} \\
            \omega_{r 1}=0 \\
            \omega_{h}=\dfrac{r_{s}}{2 r_{p 1}+2 r_{s}} \omega_{s}
            \end{array}\right. 
        &
        \left\{\begin{array}{l}
            \omega_{s}^{H}=\dfrac{r_{r 1}}{r_{r 1}+r_{s}} \omega_{s}\\
            \omega_{p 1}^{H}=\dfrac{-r_{s}\left(2 r_{p 1}+r_{s}\right)}{2 r_{p 1}\left(r_{p 1}+r_{s}\right)} \omega_{s} \\
            \omega_{r 1}^{H}=\dfrac{-r_{s}}{r_{r 1}+r_{s}} \omega_{s}\\
        \end{array}\right.
    \end{array}
    \label{eq:eq2}
\end{equation}

Substitute (\ref{eq:eq1}) into (\ref{eq:eq2}) to get the reduction ratio of the 2K-H topology as following.
\begin{equation}
    i_{2kh} = \dfrac{\omega_{s}}{\omega_{h}}=\dfrac{z_{s} + z_{r 1}}{z_{s}} 
    \label{eq:eq3}
\end{equation}

\subsubsection{Forward efficiency of 2K-H topology}

The meshing efficiency from S to P1 and P1 to R1 are $\eta_{sp1}$ and $\eta_{p1r1}$. The efficiency loss can be expressed by the loss of gear meshing tangential force $F_{r1p1}$ (R1 to P1) and $F_{p1s}$ (P1 to S). The motion state of each component in the 2K-H forward drive is shown in Fig. \ref{fig:2K-H-1} and Fig. \ref{fig:2K-H-2}, where the torque of each component in the following relationship. 
\begin{equation}
    \begin{bmatrix}
        T_{s} \\
        T_{p 1} \\
        T_{r 1}
    \end{bmatrix}=\begin{bmatrix}
        0 & r_{s} \\
        r_{p 1} & -\eta_{s p 1} r_{p 1} \\
        \eta_{p 1 r 1} r_{r1} & 0
    \end{bmatrix}\begin{bmatrix}
        F_{r 1 p 1} \\
        F_{p 1 s}
    \end{bmatrix}
    \label{eq:eq4}
\end{equation}

where,$T_{p1}=0$, substitut it in (\ref{eq:eq4}).
\begin{equation}
    \left\{\begin{aligned}
        T_{s}&=r_{s} F_{p 1 s}\\
        T_{r 1}&=\eta_{p 1 r 1} \eta_{s p 1} r_{r 1} F_{p 1 s}
    \end{aligned}\right.
    \label{eq:eq5}
\end{equation}

From Fig. \ref{fig:2K-H-2}, we can see that $T_s$ is in the same direction as $\omega_s^H$, so S is the active member. $T_{r1}$ is in the opposite direction of $\omega_{r1}^H$, and R1 is the follower. $T_{p1}=0$, P1 does not work. The meshing power loss can be found as: 
\begin{equation}
    P_{loss}=P_{s}^H - P_{r1}^H =T_{s}*\omega_{s}^H - (-T_{r1}*\omega_{r1}^H).
    \label{eq:eq6}
\end{equation}

In the stationary coordinate system, the absolute power of S is  
\begin{equation}
    P_s=T_{s}\omega_{s}.
    \label{eq:eq7}
\end{equation}

The efficiency from S to H in 2K-H is : 
\begin{equation}
    \eta_{sh}=\dfrac{P_{out}}{P_{in}}=\dfrac{P_{s}-P_{loss}}{P_{s}}
    \label{eq:eq8}
\end{equation}

Simplify $\eta_{sh}$ according to the equations above.
\begin{equation}
    \eta_{sh}=\dfrac{\eta_{sp1}\eta_{p1r1}z_{r 1} + z_{s} }{z_{r 1}+z_{s}} = \dfrac{\eta_{sp1}\eta_{p1r1}(i_{2kh}-1)+1}{i_{2kh}}
    \label{eq:eq9}
\end{equation}

\subsubsection{Backward efficiency of 2K-H topology}
The meshing efficiency from P1 to S and R1 to P1 are $\eta_{p1s}$ and $\eta_{r1p1}$. The efficiency loss can be expressed by the loss of gear meshing tangential force $F_{p1r1}$ (P1 to R1) and $F_{sp1}$ (S to P1). The motion state of each component in the 2K-H backward drive is shown in Fig. \ref{fig:2K-H-3} and Fig. \ref{fig:2K-H-4}.
\begin{equation}
    \begin{bmatrix}
    T_{s} \\
    T_{p 1} \\
    T_{r 1}
    \end{bmatrix}=\begin{bmatrix}
    0 & \eta_{p 1 s}r_{s} \\
    \eta_{r 1 p 1}r_{p 1} & - r_{p 1} \\
    r_{r} & 0
    \end{bmatrix}\begin{bmatrix}
    F_{p 1 r 1} \\
    F_{s p 1}
    \end{bmatrix}
    \label{eq:eq10}
\end{equation}

where,$T_{p1}=0$, substitut it in (\ref{eq:eq10}).
\begin{equation}
    \left\{\begin{aligned}
    T_{s}&=\eta_{r 1 p 1}\eta_{p 1 s}r_{s} F_{p 1 r 1} \\
    T_{r 1}&=r_{r} F_{p 1 r 1}
    \end{aligned}\right.
    \label{eq:eq11}
\end{equation}

From Fig. \ref{fig:2K-H-4}, we can see that $T_{r1}$ is in the same direction as $\omega_{r1}^H$, so R1 is the active member. $T_s$ is in the opposite direction of $\omega_{s}^H$, and S is the follower. $T_{p1}=0$, P1 does not work. The meshing power loss can be found as: 
\begin{equation}
    P_{loss}=P_{r1}^H-P_{s}^H = T_{r1}*\omega_{r1}^H- (-T_{s}*\omega_{s}^H)
    \label{eq:eq12}
\end{equation}

In the stationary coordinate system, the absolute power of S is
\begin{equation}
    P_s=-T_{s}\omega_{s}.
    \label{eq:eq13}
\end{equation}

The efficiency from H to S in 2K-H is 
\begin{equation}
    \eta_{hs}=\frac{P_{out}}{P_{in}}=\frac{P_{s}}{P_{s}+P_{loss}}.
    \label{eq:eq14}
\end{equation}

Simplify $\eta_{hs}$ according to the equations above.
\begin{equation}
    \eta_{hs}=\dfrac{ \eta_{r1p1} \eta_{p1s} \left(z_{r 1}+z_{s} \right) }{z_{r 1}+\eta_{r1p1} \eta_{p1s} z_{s}} = \dfrac{\eta_{r1p1} \eta_{p1s} i_{2kh}}{i_{2kh}-1+\eta_{r1p1}\eta_{p1s}}
    \label{eq:eq15}
\end{equation}

\subsection{K-H-V topology}

\subsubsection{Kinematics of K-H-V topology}

\begin{figure}[!t]
    \centering
    \subfigure[]{\includegraphics[width=0.48 \columnwidth]{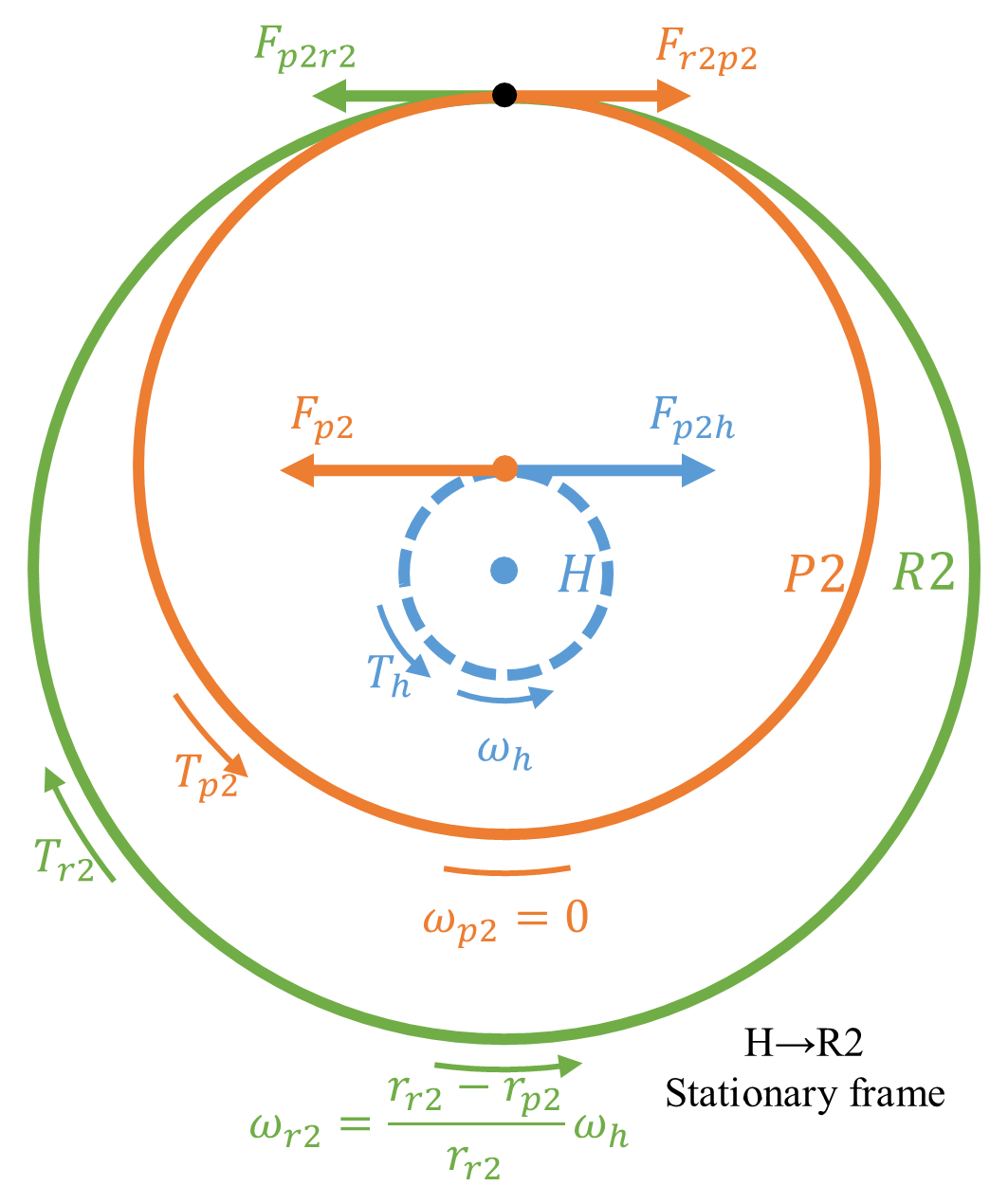}\label{fig:K-H-V-1}}%
    \hfil
    \subfigure[]{\includegraphics[width=0.48 \columnwidth]{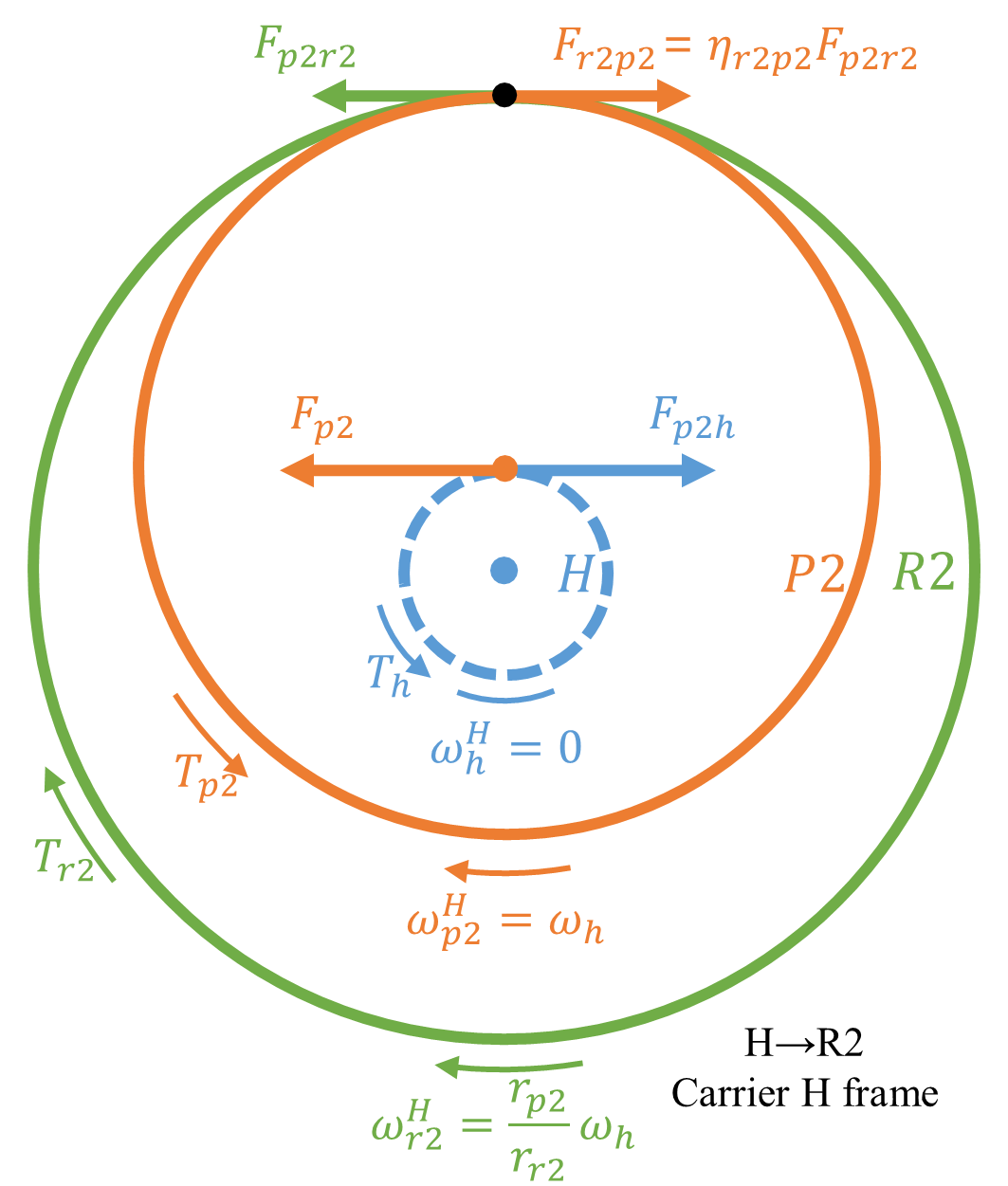}\label{fig:K-H-V-2}}%
    \\
    \subfigure[]{\includegraphics[width=0.48 \columnwidth]{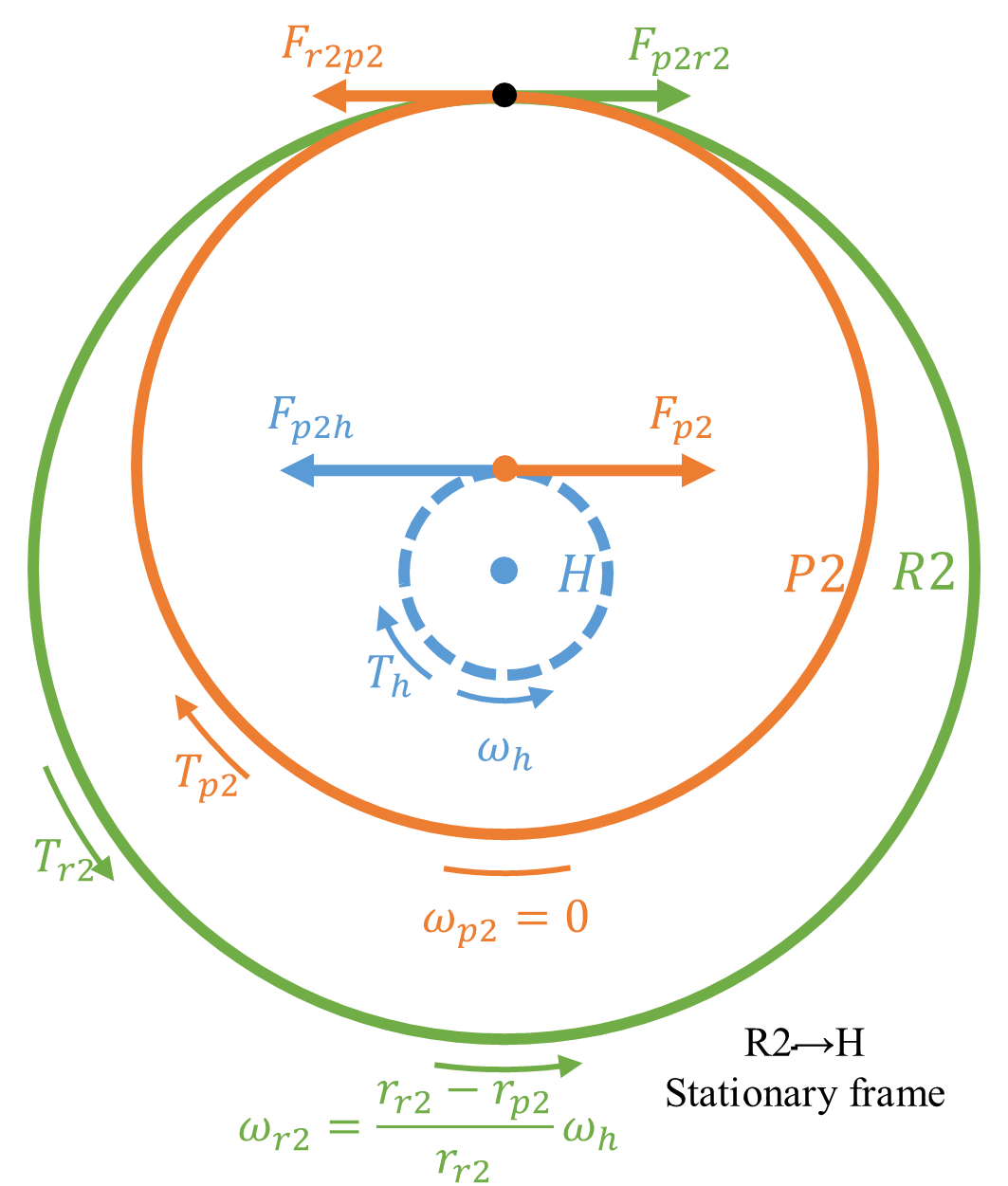}\label{fig:K-H-V-3}}%
    \hfil
    \subfigure[]{\includegraphics[width=0.48 \columnwidth]{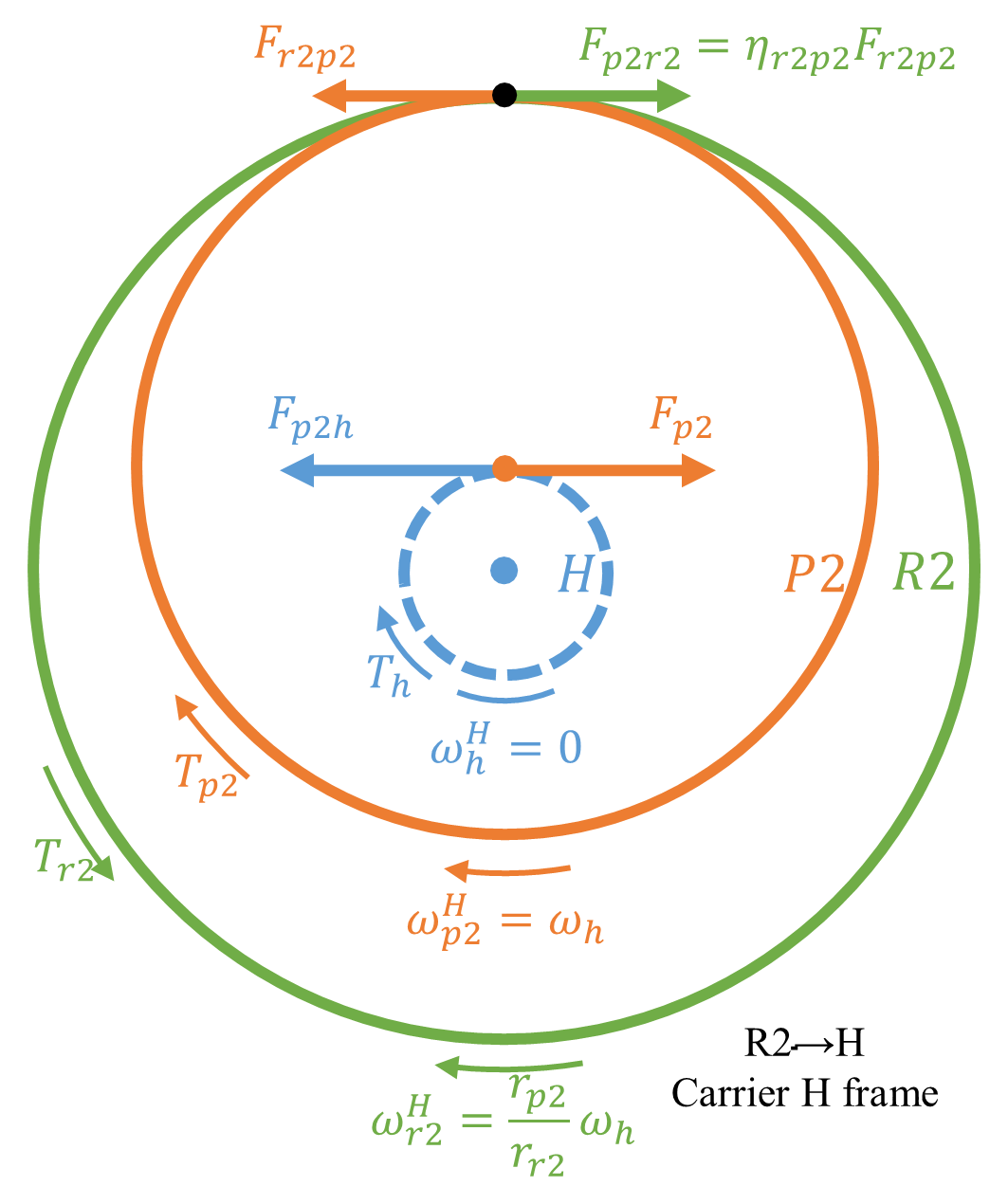}\label{fig:K-H-V-4}}%
    \caption{The motion of each component of the K-H-V topology in different conditions.}
    \label{fig:K-H-V}
\end{figure}

The K-H-V topology comprises a central gear (internal gear R2), a planetary carrier H, an output mechanism V, and planetary gears P2. The output mechanism V only transfers the spin motion of P2 as output, and the output mechanism V is not discussed in this analysis. Fig. \ref{fig:K-H-V} shows the motion of the components in K-H-V in the stationary coordinate system and H coordinate system. The drive from H to R2 is the forward drive, and the drive from R2 to H is the backward drive. 

In K-H-V, the pitch circle radiuses of R2, P2, and H are $r_{r2}$, $r_{p2}$, and $r_{h2}$, the teeth numbers of R2 and P2 are $z_{r2}$ and $z_{p2}$. The pitch circle radiuses and teeth numbers are in the following relations.
\begin{equation}
    \left\{\begin{aligned}
        \frac{r_{r2}}{z_{r2}} &= \frac{r_{p2}}{z_{p2}}\\
        r_{r2} &= r_{h2}+r_{p2} 
    \end{aligned}\right.
    \label{eq:eq16}
\end{equation}

As shown in Fig. \ref{fig:K-H-V}, in the stationary coordinate system, the absolute speed of H is $\omega_h$, the absolute speed of R2 and P2 are $\omega_{r2}$ and $\omega_{p2}$, respectively. In the coordinate system of H, the relative speeds of R2, P2, and H are $\omega_{r2}^H$, $\omega_{p2}^H$, and $\omega_{h}^H$, respectively. The absolute speeds and relative speeds are in the following relationship. The negative sign means the opposite direction.
\begin{equation}
    \begin{array}{ll}
        \left\{\begin{array}{l}
        \omega_{r 2}=\dfrac{r_{r 2}-r_{p 2}}{r_{r 2}} \omega_{h} \\
        \omega_{p 2}=0 
        \end{array}\right.
        &
        \left\{\begin{array}{l}
        \omega_{r 2}^{H}=- \omega_{h}\\
        \omega_{p 2}^{H}=\dfrac{-r_{p 2}}{r_{r 2}} \omega_{h} \\
        \omega_{h}^{H}=0 
        \end{array}\right.
    \end{array}
    \label{eq:eq17}
\end{equation}

This study uses the cycloidal gear and pin-wheel in K-H-V with the following relationships.
\begin{equation}
    z_{r 2}-z_{p 2} =1
    \label{eq:eq18}
\end{equation}

Substitute (\ref{eq:eq16}) and (\ref{eq:eq17}) into (\ref{eq:eq18}) to get the reduction ratio of the K-H-V topology as following. 
\begin{equation}
    i_{khv} = \dfrac{\omega_{h}}{\omega_{r2}}=\dfrac{z_{r 2}}{z_{r 2}-z_{p 2}} = z_{r 2}
    \label{eq:eq19}
\end{equation}

\subsubsection{Forward efficiency of K-H-V topology}
The meshing efficiency from R2 to P2 is $\eta_{r2p2}$. The efficiency loss can be expressed by the loss of gear meshing tangential force $F_{p2r2}$ (P2 to R2). The motion state of each component in the K-H-V forward drive is shown in Fig. \ref{fig:K-H-V-1} and Fig. \ref{fig:K-H-V-2}, where the torque of each component in the following relationship. 
\begin{equation}
    \left\{\begin{aligned}
        T_{r 2}&=r_{r2} F_{p 2 r 2}\\
        T_{p 2}&=-\eta_{r 2 p 2}r_{p2} F_{p 2 r 2}
    \end{aligned}\right.
    \label{eq:eq20}
\end{equation}

From Fig. \ref{fig:K-H-V-2}, we can see that $T_{r 2}$ is in the same direction as $\omega_{r2}^H$, so R2 is the active member. $T_{p2}$ is in the opposite direction of $\omega_{p2}^H$, and P2 is the follower. The meshing power loss can be found as 
\begin{equation}
    P_{loss}=P_{r2}^H-P_{p2}^H = T_{r2}*\omega_{r2}^H - (-T_{p2}*\omega_{p2}^H).
    \label{eq:eq21}
\end{equation}

In the stationary coordinate system, the absolute power of R2 is 
\begin{equation}
    P_{r2}=-T_{r2}\omega_{r2}.
    \label{eq:eq22}
\end{equation}

The efficiency from H to R2 in K-H-V is
\begin{equation}
    \eta_{hr2}=\frac{P_{out}}{P_{in}}=\frac{P_{r2}}{P_{r2}+P_{loss}}.
    \label{eq:eq23}
\end{equation}

Simplify $\eta_{hr2}$ according to the equations above.
\begin{equation}
    \eta_{hr2}=\dfrac{z_{r 2}-z_{p 2}}{z_{r 2}-\eta_{r2p2}z_{p 2}} = \dfrac{1}{\eta_{r2p2} + i_{khv}(1 - \eta_{r2p2})} 
    \label{eq:eq24}
\end{equation}

\subsubsection{Backward efficiency of K-H-V topology}
The meshing efficiency from P2 to R2 is $\eta_{p2r2}$. The efficiency loss can be expressed by the loss of gear meshing tangential force $F_{r2p2}$ (R2 to P2). The motion state of each component in the K-H-V backward drive is shown in Fig. \ref{fig:K-H-V-3} and Fig. \ref{fig:K-H-V-4}.
\begin{equation}
    \left\{\begin{aligned}
    T_{r 2}&=-\eta_{p 2 r 2}r_{r2} F_{r 2 p 2} \\
    T_{p 2}&=r_{p2} F_{r 2 p 2}
    \end{aligned}\right.
    \label{eq:eq25}
\end{equation}

From Fig. \ref{fig:K-H-V-4}, we can see that $T_{p2}$ is in the same direction as $\omega_{p2}^H$, so P2 is the active member. $T_{r2}$ is in the opposite direction of $\omega_{r2}^H$, and R2 is the follower. The meshing power loss can be found as
\begin{equation}
    P_{loss}=P_{p2}^H-P_{r2}^H = T_{p2}*\omega_{p2}^H - (-T_{r2}*\omega_{r2}^H).
    \label{eq:eq26}
\end{equation}

In the stationary coordinate system, the absolute power of R2 is 
\begin{equation}
    P_{r2}=T_{r2}\omega_{r2}.
    \label{eq:eq27}
\end{equation}

The efficiency from R2 to H in K-H-V is
\begin{equation}
    \eta_{r2h}=\frac{P_{out}}{P_{in}}=\frac{P_{r2}-P_{loss}}{P_{r2}}.
    \label{eq:eq28}
\end{equation}

Simplify $\eta_{r2h}$ according to the equations above.
\begin{equation}
  \eta_{r2h}=\dfrac{\eta_{p2r2}z_{r 2}-z_{p 2}}{\eta_{p2r2}\left(z_{r 2}-z_{p 2}\right)} = \dfrac{1+i_{khv}(\eta_{p2r2}-1)}{\eta_{p2r2}}
  \label{eq:eq29}
\end{equation}

\begin{figure}[!t]
    \centering
    \subfigure[]{\includegraphics[width=0.9 \columnwidth]{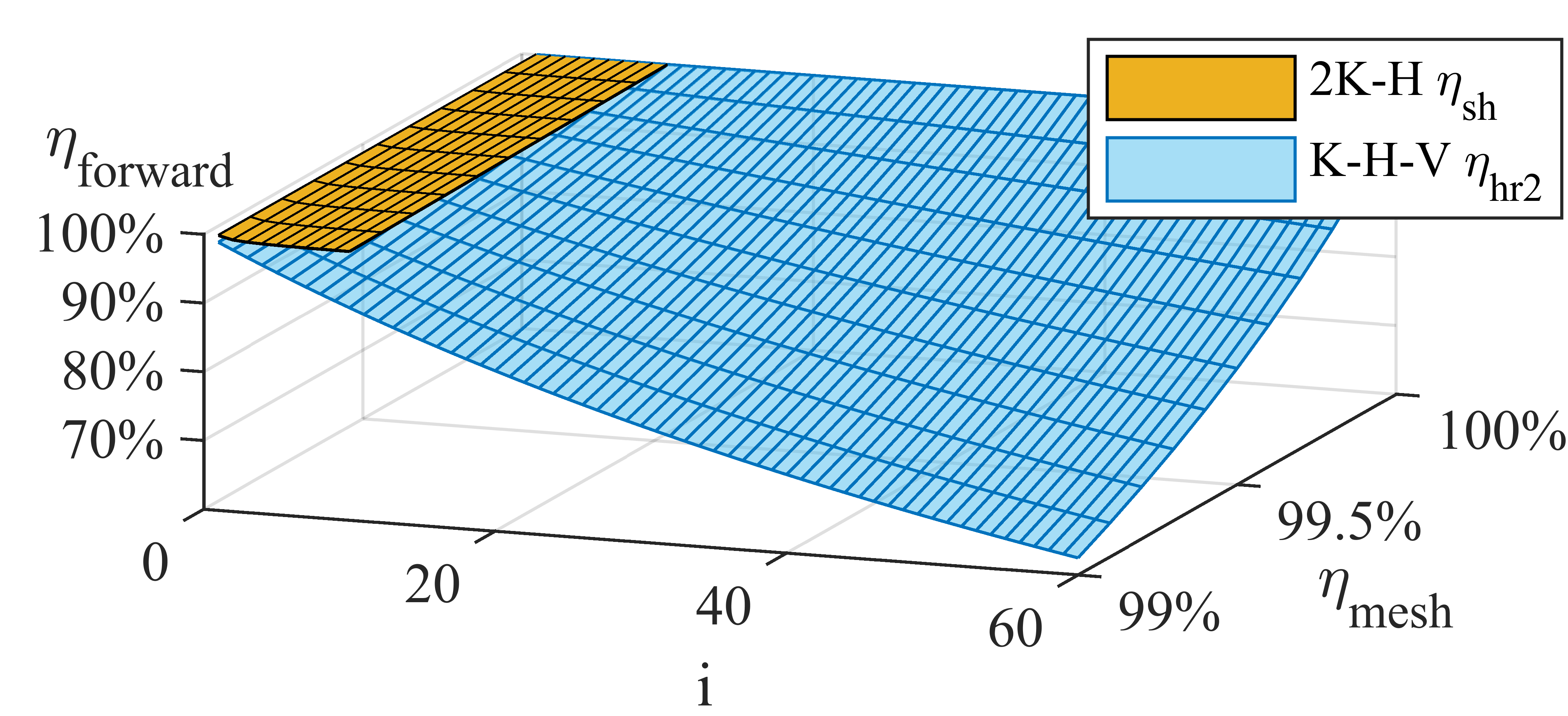}\label{fig:eff-1}}\\
    \subfigure[]{\includegraphics[width=0.9 \columnwidth]{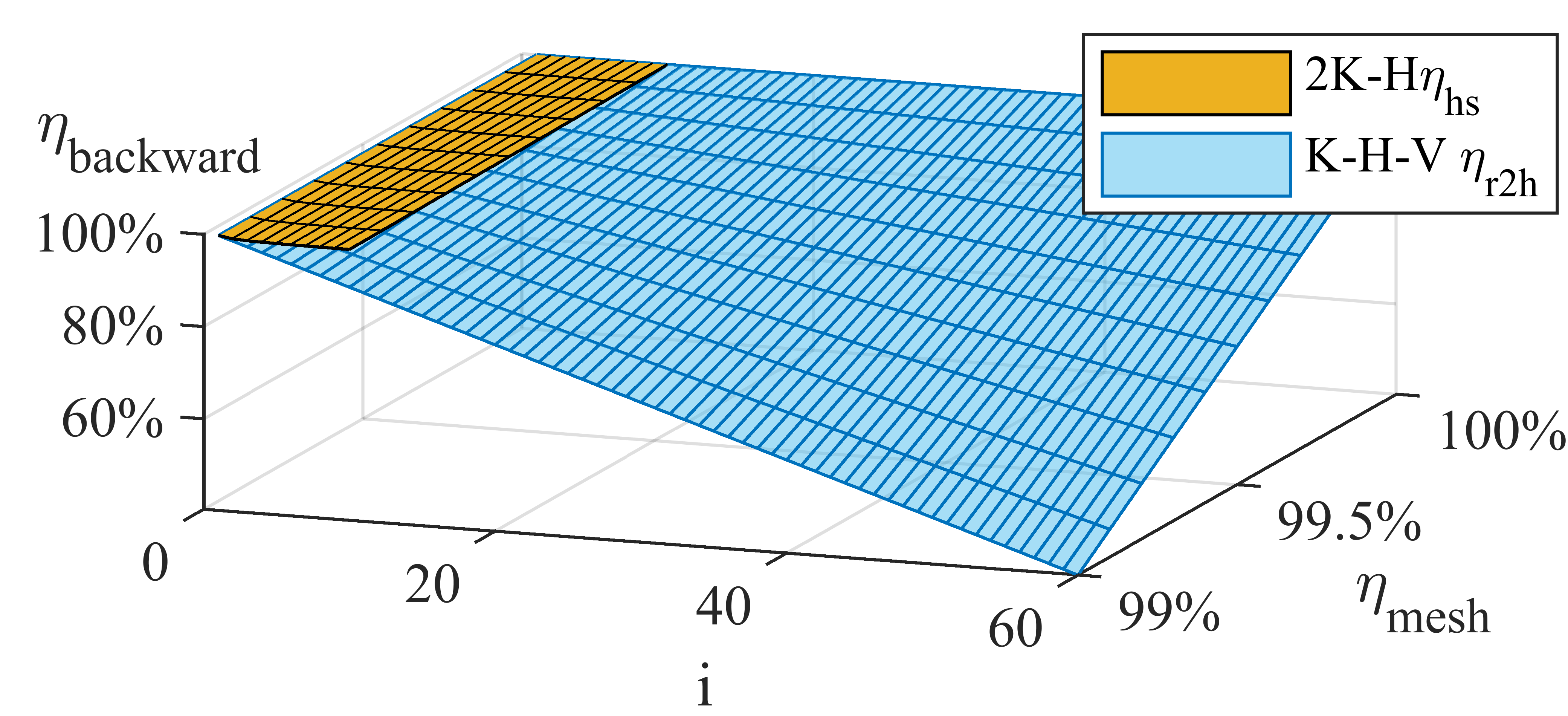}\label{fig:eff-2}}%
    \caption{(a) The forward and (b) backward efficiency of 2K-H and K-H-V}
    \label{fig:eff}
\end{figure}

\subsection{3K-H-V topology}

The 2K-H and K-H-V stages have the same planetary carrier H in the 3K-H-V. 

The reduction ratio of 3K-H-V is
\begin{equation}
    i_{3khv}=i_{2kh} i_{khv} = \dfrac{z_{r 2}(z_s+z_{r1})}{z_{s}}.
    \label{eq:eq30}
\end{equation}

The forward and backward efficiency of 3K-H-V are
\begin{equation}
    \left\{\begin{aligned}
        \eta_{sr2}=\eta_{sh}\eta_{hr2}\\
        \eta_{r2s}=\eta_{r2h} \eta_{hs}
    \end{aligned}\right..
    \label{eq:eq31}
\end{equation}

The reduction ratio of 2K-H usually does not over 10, and the meshing efficiency is generally greater than 99\%. In the two stages of 3K-H-V, with the increase of the reduction ratio and the decrease of the meshing efficiency $\eta_{mesh}$ (i.e., $\eta_{sp1}\eta_{p1r1}$, $\eta_{r2p2}$, $\eta_{r1p1}\eta_{p1s}$ and $\eta_{p2r2}$), both forward and reverse efficiencies will decrease, and the backward efficiency is lower than the forward efficiency, as shown in Fig. \ref{fig:eff}. 

In the design of PCD, the overall efficiency can be optimized by increasing the percentage of the 2K-H reduction ratio to the total reduction ratio while achieving the target reduction ratio. In addition, a meaningful way to improve the overall efficiency of the PCD is to improve the meshing efficiency of each group of gears, which will be an essential direction for later research. It is worth noting that when $\eta_{p2r2}\le \frac{z_{p 2}}{z_{r 2}}$, the backward efficiency of K-H-V decreases to non-positive (self-locking), which can be applied to some unique applications but needs to be avoided in most applications.

  \section{EXPERIMENTAL RESULTS OF THE PCA PROTOTYPE}
\begin{figure}[!htbp]
    \centering
    \includegraphics[width=\columnwidth]{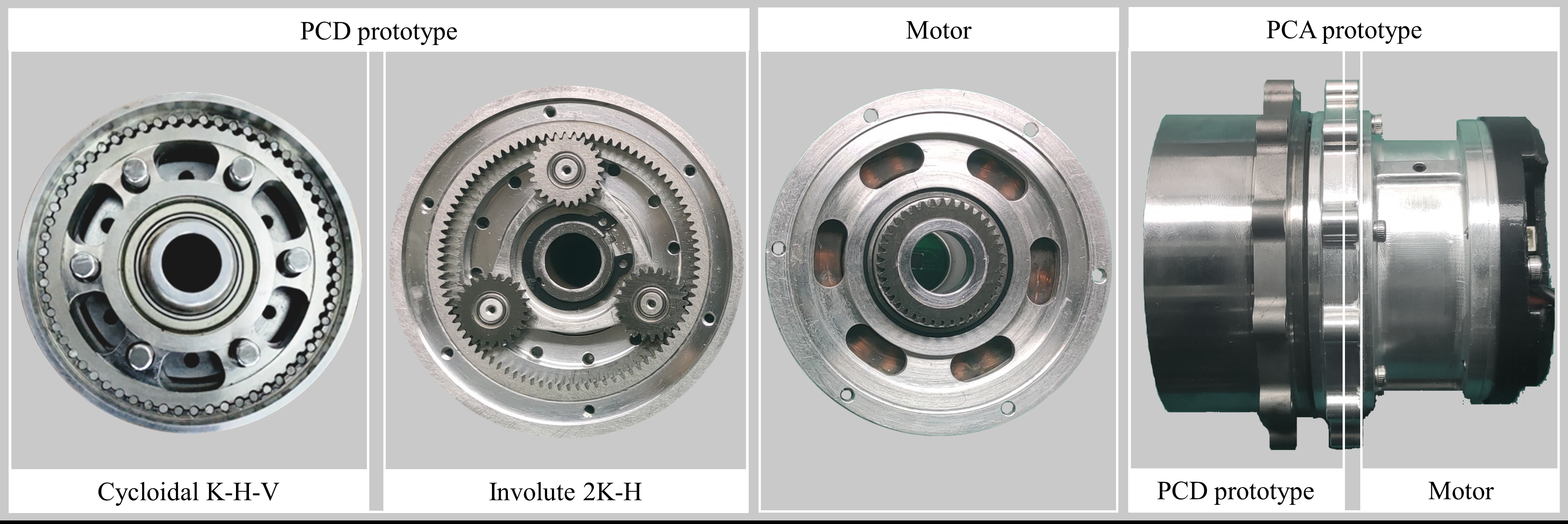}
    \caption{PCA prototype}
    \label{fig:3khv-1}
\end{figure}

\begin{figure}[!htbp]
    \centering
    \includegraphics[width=\columnwidth]{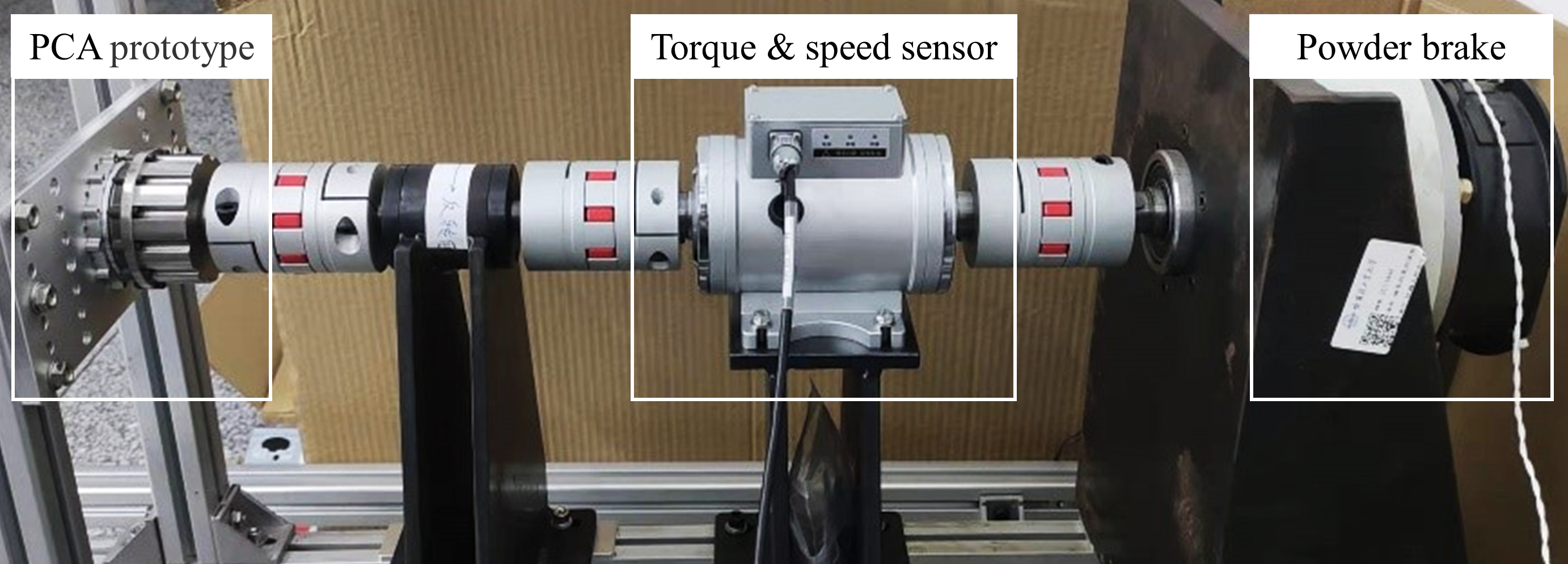}
    \caption{Experimental setup}
    \label{fig:3khv-4}
\end{figure}

The PCA prototype equipped with the PCD drive based on 3K-H-V has been completed, as shown in Fig. \ref{fig:3khv-1}. The experimental setup is shown in Fig. \ref{fig:3khv-4}. 

\begin{table}[!h]
    \centering
    \caption{Design parameters and experimental results}
    \label{table:pcd}
    \begin{tabular}{lc}
        \hline
        \multicolumn{2}{c}{Design parameters}   \\ \hline
        Teeth number & \begin{tabular}[c]{@{}c@{}}
            $z_{s} = 39$, $z_{p1} = 24$, $z_{r1} = 87$\\      
            $z_{p2} = 59$, $z_{r2} = 60$
        \end{tabular} \\ 
        Gear ratio   & $i_{2kh} (3.23) \times i_{khv} (60) = i_{3khv} (193.8)$  \\ 
        Size    & \multicolumn{1}{c}{$\Phi 82 mm \times 80 mm$ with hollow $\Phi 12 mm$} \\ \hline
        \multicolumn{2}{c}{Experimental results} \\ \hline
        Mass    & $\text{PCD} (0.65 kg) + \text{Motor} (0.26 kg) = \text{PCA} (0.91 kg)$ \\ 
        Rated torque   & 63 Nm   \\ 
        \begin{tabular}[l]{@{}l@{}}Rated torque density 
        \end{tabular}& 69 kg/Nm\\
        PCD efficiency     & 65\% (Forward)  \\ \hline
    \end{tabular}
\end{table}

From the experimental results, the PCA prototype achieves a well rated torque in a lightweight and compact size with a hollow structure. It has a high rated torque density of (69 Nm/kg). The PCD prototype can achieve a high reduction ratio of $193.8$. However, the PCD prototype has a low forward drive efficiency of $(65\%)$, which may be related to the low meshing efficiency due to machining errors. The PCA prototype initially verifies the feasibility of the PCD for the robot actuator.

\section{CONCLUSIONS AND FUTURE WORK}
This study proposes a planetary cycloidal drive based on 3K-H-V by combining the advantages of 2K-H and K-H-V topologies with the advantages of involute and cycloidal through the design idea of decoupling to meet the needs of robot actuator with high precision, high stiffness, high load, high efficiency, low backlash, and hollow structure. This study analyzes the reduction ratio and forward and reverse drive efficiency of 3K-H-V and proposes ideas to improve the efficiency. This study designs a prototype robot actuator on this principle of drive and conducts a preliminary experimental verification of the load carrying and efficiency.

The planetary cycloidal drive based on 3K-H-V proposed in this paper theoretically takes the advantages of high precision, high stiffness, high loading, and low backlash of the cycloidal K-H-V. It uses the involute 2K-H to improve the reduction ratio of the drive and ensure high efficiency. The actuator prototype PCA with PCD verifies the advantages of high torque density (69 Nm/kg), high reduction ratio, compact size, and hollow structure. However, there is still a problem of low efficiency.

In future research, we will further test the accuracy and stiffness characteristics of the PCD and model and analyze the meshing efficiency to propose a method to improve the PCD efficiency.

\addtolength{\textheight}{0cm}

\bibliographystyle{IEEEtran}
\bibliography{reducer-zotero.bib}
\end{document}